\documentclass[wcp]{jmlr}
\usepackage{amsmath}
\usepackage{amssymb}
\usepackage{multirow}
\usepackage{rotating}
\usepackage{amsbsy}
\usepackage{cases}
\usepackage{amscd}
\usepackage[normalem]{ulem}
\usepackage{latexsym}
\usepackage{ bbold }
\usepackage{ mathrsfs }
\usepackage[T1]{fontenc}
\usepackage{relsize}
\usepackage[english]{babel}
\usepackage{ dsfont }
\usepackage{graphicx}
\usepackage{xpatch}

\usepackage{ifthen}
\usepackage{algorithm}
\usepackage{algorithm,algpseudocode}% http://ctan.org/pkg/{algorithms,algorithmx}
\usepackage{forest}
\allowdisplaybreaks
\usetikzlibrary{arrows.meta}
\algnewcommand{\Inputs}[1]{%
	\State \textbf{Inputs:}
	\Statex \hspace*{\algorithmicindent}\parbox[t]{.8\linewidth}{\raggedright #1}
}
\algnewcommand{\Initialize}[1]{%
	\State \textbf{Initialize:}
	\Statex \hspace*{\algorithmicindent}\parbox[t]{.8\linewidth}{\raggedright #1}
}
\usepackage{longtable}% for long tables

\usepackage{booktabs}

\DeclareMathOperator*{\argmax}{arg\,max}
\DeclareMathOperator*{\depth}{depth}

\DeclareMathOperator*{\kl}{kl}
\DeclareMathOperator*{\supp}{supp}
\DeclareMathOperator*{\h}{H}
\DeclareMathOperator*{\T}{T}
\DeclareMathOperator*{\bro}{Br-o}
\DeclareMathOperator*{\To}{T-o}
\DeclareMathOperator*{\br}{Br}
\DeclareMathOperator*{\Cr}{Cr}
\DeclareMathOperator*{\StoROO}{StoROO}
\DeclareMathOperator*{\klp}{kl-p}
\DeclareMathOperator*{\ber}{B}
\DeclareMathOperator*{\Lip}{Lip}
\DeclareMathOperator*{\diam}{diam}
\DeclareMathOperator*{\cvar}{CVaR_{\tau}}
\DeclareMathOperator*{\bet}{\hat{\beta}}
\DeclareMathOperator*{\gamm}{\hat{\gamma}}
\jmlrvolume{101}
\jmlryear{2019}
\jmlrworkshop{ACML 2019}

\title[Stochastic Risk Optimistic Optimization]{$\mathcal{X}$-Armed Bandits: Optimizing Quantiles, CVaR and Other Risks}

 \author{\Name{L\'eonard Torossian} \Email{leonard.torossian@inra.fr}\\
  \addr Universite de Toulouse, INRA, France \and
  Institut de Mathematiques de Toulouse, France
  \AND
  \Name{Aur\'elien Garivier} \Email{aurelien.garivier@ens-lyon.fr}\\
  \addr Univ. Lyon, ENS de Lyon, France
  \AND
  \Name{Victor Picheny} \Email{victor@prowler.io}\\
  \addr PROWLER.io, 72 Hills Road, Cambridge, UK
 }

\editors{Wee Sun Lee and Taiji Suzuki}

\begin{document}

\maketitle

\begin{abstract}
We propose and analyze StoROO, an algorithm for risk optimization on stochastic black-box functions derived from StoOO.
Motivated by risk-averse decision making fields like agriculture,
medicine, biology or finance, we do not focus on the mean payoff but on
generic functionals of the return distribution.
We provide a generic regret analysis of StoROO and illustrate its applicability with two examples: the optimization of quantiles and CVaR.
Inspired by the bandit literature and black-box mean optimizers,
% 	like StoSOO and HCT, 
StoROO relies on the possibility to construct confidence intervals
for the targeted functional based on random-size samples. We detail their construction in the case of quantiles, providing tight bounds based on Kullback-Leibler divergence. We finally present numerical experiments that show a dramatic impact of tight bounds for the optimization of quantiles and CVaR.
%these bounds and illustrate the ability of StoROO to optimize the CVaR.
% 	so far known methods \leonard{on a le droit de dire cette phrase sans avoir teste la version bernstein ?}.
\end{abstract}

	\section{Introduction}
	We consider an unknown function $\Phi : \mathcal{X} \times \Omega \rightarrow [0,1] \subset \mathds{R}$, where $\mathcal{X}\subset[0,1]^D$ and $\Omega$ denotes the probability space representing some uncontrollable variables. 
	For any fixed $x\in \mathcal{X}, Y_x = \Phi(x, \cdot)$ is a random variable of distribution $\mathds{P}_x$ and we consider $g(x) = \psi(\mathds{P}_x)$ with $\psi$ a real-valued functional defined on probability measures. 
	We assume that	there exists at least one $x^*\in\mathcal{X}$ such that $g(x^*)=\sup_{x\in\mathcal{X}}g(x)$. Using a set of sequential observations $(\Phi(x_1,\omega_1),\cdots,\Phi(x_T,\omega_T))$, our goal is to minimizing the simple regret 
	$r_T=g(x^*)-g(x_{T}),$
	with $x_{T}$ the value returned after using a budget $T$.
	
	Different families of algorithms have been developed to treat this problem. Some are for example of Bayesian flavor \citep[see ][for instance]{shahriari2016taking}, some are inspired by the bandit literature. Here we focus our interest on the bandit framework.
	
	In the classical $\mathcal{X}$-armed bandit problem, a forecaster selects repeatedly a point $x$ in the input space $\mathcal{X}\in[0,1]^D$ and receives a reward distributed according to an unknown distribution $\mathds{P}_x$. 
	 Historically, the main goal was to minimizing the \textit{cumulative regret}, i.e. the sum of the difference between his collected rewards and the ones that would have been brought by optimal actions. In the last decade, other works focused on the simple regret. These can be divided in two: algorithms that optimize an unknown function with the knowledge of the  smoothness, 
	 for example StoOO \citep{munos2014bandits}, HOO \citep{bubeck2011x} or Zooming \citep{kleinberg2008multi} and others focusing on the optimization
	 of unknown functions without the knowledge of the smoothness, such as POO \citep{grill2015black}, StroquOOL \citep{bartlett2018simple}, GPO \citep{xuedong2019general}, StoSOO \citep{valko2013stochastic} or \citet{locatelli2018adaptivity}.

	Those algorithms focus on the optimization of the conditional \emph{expectation} of $\mathds{P}_x$. This choice is questionable in some situations. 
	For example if the shape and variance of the reward distribution depend on the input, a forecaster may be interested in different aspects of the unknown distribution in order to modulate its risk exposure.
	In the literature, some measures of risk have been proposed to replace the expectation: for instance quantiles \citep[also referred to as Value-at-Risk, see ][]{artzner1999coherent},
	the Conditional Value-at-Risk \citep[CVaR also referred as Superquantile or Expected Shortfall, ][]{rockafellar2000optimization} or expectiles \citep{bellini2017risk}.
	The purpose of this paper is to present a risk optimization framework of an unknown stochastic function with the knowledge of the smoothness using only pointwise sequential observations and a finite budget $T$. 
	
	$\mathcal{X}$-armed bandit algorithms rely on \textit{optimistic strategies} that associate with each point of the space an upper confidence bound (UCB), that is, an \textit{optimistic} prediction of the outcome. 
	Adapting the classical setting to the optimization of risk measures implies being able to create high-probability confidence bounds for that particular measure. 
	This problem has been tackled in the multi-armed bandit setting ($i.e.$ when the input space is discrete and finite). 
	For instance, \citet{audibert2009exploration,sani2012risk} focused on the empirical variance, \citet{galichet2013exploration,kolla2019risk,hepworth2017multi} on the CVaR 
	while in \citet{david2016pure,szorenyi2015qualitative} the authors based their policies on the quantile. 
	However, the literature is scarce in the continuous input space case.
	
	In this paper we provide a new version of the Stochastic Optimistic Optimization (StoOO) algorithm \citep{munos2014bandits}, named StoROO (Stochastic Risk Optimistic Optimization), which is
	designed to handle any functional $\psi$.
	In a first part, we provide an analysis of the simple regret from a generic point of view.
	We then particularize our analysis in two important illustrative cases: conditional quantiles and CVaR. 
	In the case of quantiles, assuming that the output distribution has a continuous, strictly increasing cumulative distribution function, we first propose an upper bound on the simple regret using Hoeffding's inequality, then, we derive tighter confidence intervals that take into account the order of the quantile respectively based on Bernstein's and Chernoff's inequalities. In the case of the CVaR, we first derive an upper bound on the regret using the deviation inequality of \cite{brown2007large}, then using the work of \cite{thomas2019concentration} we derived tighter confidence bounds.  %\victor{something is missing to announce what we do about the CVaR}
	Finally, we present numerical experiments that illustrate the ability of our method to optimize conditional quantiles and CVaR of a black-box function and the relevance of using tight deviation bounds. 
	\section{Problem setup}
	\subsection{Hierarchical partitioning}
% 		\textit{Optimistic algorithms} rely on an upper confidence bound (UCB) of the objective function instead of a simple  estimator, that is, 
	The upper confidence bounds on which optimistic algorithms are based are surrogate functions 
	$U:\mathcal{X}\rightarrow \mathds{R}$ larger than the objective (in a sense detailed below) with high probability.
	At each round $t$, the point $X(t)$ having the highest UCB is sampled and a reward $Y_X(t)$ is collected. 
	
	In the classical multi-armed bandit problem, computing and sorting the UCB can be done without major issues. 
	But dealing with continuous input spaces implies maximizing a UCB function over a continuous space, which can be both computational intensive
	and algorithmically challenging. 
% 	computing and 
	For example, Piyavskii's algorithm \citep[see][and references therein]{bouttier2017optimisation} defines $U$ using a global Lipschitz assumption on the targeted function. 
	Because of the Lipschitz hypothesis, the UCB maximizer is at an intersection of hyperplanes, i.e. where the UCB is non-differentiable. 
	Thus a gradient-based algorithm cannot be used, implying that finding the point with the highest UCB is a very hard problem to solve.
	
	To overcome the computational difficulties, a popular alternative is to rely on hierarchical partitions (see \cite{bubeck2011x,munos2014bandits} for instance), $\mathcal{P}=\{\mathcal{P}_{h,j}\}_{h,j}$ of $\mathcal{X}$ such that
	\begin{equation*}
	 \mathcal{P}_{0,1}=\mathcal{X},~~\mathcal{P}_{h,j}=\bigcup_{i=0}^{K-1}\mathcal{P}_{h+1,Kj-i}\;,
	\end{equation*}
with $K$ the number of sub-regions obtained after expanding a cell and $\mathcal{P}_{h,j}$ the $j$-th cell at depth $h$. In the following we assume that:

	\textbf{Assumption $1$:}
	 There exists a decreasing sequence $\delta(h)$, such that for any $h\geq 0$ and for any cell $\mathcal{P}_{h,j}$, 
	$\sup_{x\in\mathcal{P}_{h,j}} \left\| x-x_{h,j}\right\| _{\infty} \leq \delta(h)$, with $x_{h,j}$ the center of $\mathcal{P}_{h,j}$.

	\textbf{Assumption $2$:}
	There exists $\nu>0$ such that every cell of depth $h$ contains a ball of radius $\nu \delta(h)$.
\iffalse
	Different kind of hierarchical space structure can be consider. The simple approach consider the division of the cells into hyper-rectangles. That can be done by splitting a cell into $K$ children only through one dimension. A second case that will be what we considere here is 	the sub-regions are created by the division of the parent-cell into $K=2^D$ sub-regions of equal size.
	\fi
	
	Starting with $\mathcal{P}_{0,1}$ and following an optimistic strategy, at time $t$ the algorithm has expanded some cells and the result is a tree $\mathcal{T}_t$ that is a subset of $\mathcal{P}$ and a partition of $\mathcal{X}$. 
	In this setting $U$ is taken as a piecewise constant function. Indeed for any $(\mathcal{P}_{h,j})_{h,j\in\mathcal{T}_t}$ we define $\bar{U}_{h,j}$ such that for all $x\in\mathcal{P}_{h,j}$, $U(x)=\bar{U}_{h,j}$. 

	In the literature of $\mathcal{X}$-armed bandits there are two ways to select a cell of $\mathcal{T}_t$ at each round. 
	In \cite{bubeck2011x}, the algorithm follows an \textit{optimistic path} from the root to the leaves.
	In \cite{munos2014bandits}, StoOO selects the cell having the highest UCB among all the cells of $\mathcal{T}_t$ that have not been expanded, $i.e.$ the set $\mathcal{L}_t$ of leaves of $\mathcal{T}_t$. 
	We consider here this second alternative. Hence, to find the maximizer of $U$ at time $t$, we only need to evaluate and sort a finite number of values $(\bar{U}_{h,j})_{(h,j)\in\mathcal{L}_t}$.
	
\subsection{Regularity assumptions, noise and bias}
%The algorithms considered in this paper rely on the possibility to upper-bound with high probability the value of the function in a cell based on a sample. This requires to control two terms: a bias term related to the variations of the objective on the cell, and a variance term related to the level of noise. 
%\victor{Rework the transition? and start directly with the regularity assumption?}

Even in the absence of noise, optimization from finite samples requires some regularity of the objective. Following \cite{munos2014bandits}, we assume the following smoothness property:
	\begin{equation}\label{reg}
\forall  x\in\mathcal{X}, \quad g(x)\geq g(x^*)-\beta||x-x^*||^\gamma \text{ with }\gamma,~\beta~> 0\;.
	\end{equation}
	Note that this condition is less restrictive than a global H\"older condition. In particular, the objective may be very irregular (even possibly discontinuous) except in the neighborhood of global maxima. 
	
	At first glance, in our stochastic setting, it may not be easy to asses that $g$ satisfies (\ref{reg}). 
Sufficient conditions can be derived from the continuity of the conditional distribution $\mathds{P}_x$ with respect to $x$. %\victor{is that true for any functional? I don't think so. If not, change to ``(...) can be derived for the quantile and CVaR for instance (...)''}. In the case of parametric $\mathds{P}_x$, it generally means that the parameters must satisfy a condition similar to \eqref{reg}.
The relevant metric on the space of distributions actually depends on the chosen risk. For conditional quantiles, the natural assumption is that $x\mapsto F_x^{-1}(\tau)$ satisfies (\ref{reg}), and a sufficient condition is that $\|F_x^{-1}-F_y^{-1}\|_\infty \leq \beta \|x-y\|^\gamma$. In the case of the the conditional expectation and for the CVaR (or more generally for a large class of Optimized Certainty Equivalent \cite{ben2007old}), the natural metric involved is the \emph{Wasserstein distance} $\mathcal{W}_1$, as explained in Section \ref{apd:details}. 

To create confidence bounds for $(\mathcal{P}_{h,j})_{(h,j)\in\mathcal{L}_t}$, StoOO samples the leafs at their centers $(x_{h,j})_{(h,j)\in\mathcal{L}_t}$. Then
using that all observed values are independent, \emph{deviation inequalities} are used to create $(U_{h,j})_{(h,j)\in\mathcal{L}_t}$, a UCB for $(g(x_{h,j}))_{(h,j)\in\mathcal{L}_t}$. 
Finally to create $(\bar{U}_{h,j})_{(h,j)\in\mathcal{L}_t}$, a UCB over the cells, a \emph{bias term} is added that takes into account how $g$ can potentially increase from the center of the cell to its edges. Because the convergence of StoOO (and StoROO) only needs $\bar{U}_{h,j}$ to be a UCB of $\max_{x\in\mathcal{P}_{h,j}}g(x)$ for the cell containing $x^*$ 
(see the proof of Proposition~\ref{expand} (see also~\cite{munos2014bandits}), it is enough to use the condition (\ref{reg}) to define a UCB as
		\begin{equation*}
	  \bar{U}_{h,j}=U_{h,j}+ B_{h,j}, \text{ with } B_{h,j} = \hat{\beta}  \delta(h)^{\hat{\gamma}}\;,
	\end{equation*}
	$\text{and}~\beta\leq\hat{\beta},~\gamma\geq\hat{\gamma}$. The algorithm also needs a quantity that bounds $g$ from below in order to provide guaranties on the value of $g$ over each cell. We thus construct a lower confidence bound, termed $L_{h,j}$, for $g(x_{h,j})$, and use it as a LCB for the maximum of $g$ on $\mathcal{P}_{h,j}$.
	In particular, on the cell $\mathcal{P}_{h^*,j^*}$ containing the optimum $x^*$, it holds that
	\begin{equation*}
	  L_{h^*,j^*} \leq g(x^*)\leq U_{h^*,j^*}+ \hat{\beta}  \delta(h^*)^{\hat{\gamma}}\;
	\end{equation*}
with high probability.
	To summarize, the estimation of $g(x^*)$ is altered by two sources of error: 
	the local estimation error $E_{h^*,j^*} = U_{h^*,j^*}-L_{h^*,j^*}$ made at the center of the cell, and the bias term $B_{h^*,j^*}$.
	Balancing those two terms naturally provides a trade-off between exploration and exploitation.
	
% 	In the following we define 
% 	\begin{equation*}
% 	  E_{h,j}=U_{h,j}-L_{h,j}~~\text{and}~~B_{h,j}=\beta\delta(h)^\gamma.
% 	  \end{equation*}
% 	Associating this two contributions, we define  $\bar{E}_{h,j}=E_{h,j}+B_{h,j}$. 

	\section{Stochastic Risk Optimistic Optimization}

\subsection{The StoROO algorithm}

StoROO starts by sampling one time each $K$ sub-region of the root node. 
	Then, at each time $1\leq t \leq T$ the algorithm selects $\mathcal{P}_{h_t,j_t}\in(\mathcal{P}_{h,j})_{(h,j)\in\mathcal{L}_t}$ having the highest UCB. 
	To reduce the estimation error, StoROO can either get more samples from $\mathcal{P}_{h_t,j_t}$ (to reduce the variance), or split the cell in order to reduce its diameter (to reduce the bias). 
	The good balance between these two options is found by dividing a cell as soon as the local estimation error is smaller than the bias, that is when
	\begin{equation}\label{explose}
	U_{h_t,j_t}-L_{h_t,j_t}\leq \hat{\beta}  \delta(h_t)^{\hat{\gamma}}\;.
	\end{equation}
	If Condition (\ref{explose}) is satisfied, StoROO expands $\mathcal{P}_{h_t,j_t}$ and requires a new sample at the center of each sub-region. 
	If Condition (\ref{explose}) is not satisfied, then StoROO requires a new sample at the center $x_{h_t,j_t}$ which is used to update $U_{h_t,j_t}$ and $L_{h_t,j_t}$.
	
	When the budget is exhausted, several choices are possible for the return value: they have the same theoretical guarantees. 
	Following \cite{munos2014bandits}, one can return the deepest node among those that have been expanded. 
	Here we propose a different, more conservative choice. Denoting by $\mathscr{L}_T$ the set of nodes having the highest LCB among those that have been expanded after a budget $T$, 
	StoROO returns the node with the highest value $\hat{g}$ (an estimator of $g$) among the deepest nodes of $\mathscr{L}_T$. The pseudo-code of the full algorithm is given in Algorithm~\ref{sto}.
It requires the parameters $\hat{\beta}$ and $\hat{\gamma}$ that satisfy Condition (\ref{reg}), but of course the inequality do not have to be tight.

	\begin{algorithm}
	\KwIn{error probability $\eta>0$; number of children $K$; time horizon $T$; $\hat{\beta}>0$; $\hat{\gamma}>0$;}
	\textbf{Define:} UCB and LCB% of $g$; $\hat{g}$\;
	
\textbf{Initialization}	$n=1$; $t=1$;\;

 Expand into $K$ sub-regions the root node $(0,0)$ and sample one time each child\;

\While{$n$ $\leq$ $T$}{

	\ForEach{ $(h,j)\in \mathcal{L}_t$ }{
		compute $\bar{U}_{h,j}(t)$\;
	}
	 Select $(\tilde{h},\tilde{j})=\argmax_{(h,j)\in\mathcal{L}_t} \bar{U}_{h,j}(t)$\;
	 
	 Compute the LCB $L_{\tilde{h},\tilde{j}}(t)$\;
	 
	 \eIf{$U_{\tilde{h},\tilde{j}}(t)-L_{\tilde{h},\tilde{j}}(t)\leq  \hat{\beta} \delta(h)^{\hat{\gamma}}$}{
		expand the node,
				remove $(\tilde{h},\tilde{j})$ from $\mathcal{L}_t$, add to $\mathcal{L}_t$ the $K$ sub-cells of $\mathcal{P}_{\tilde{h},\tilde{j}}$ and sample each new node once,
				
				$n=n+K$, $t=t+1$\;
	}
	{Sample the state $x_t=x_{\tilde{h},\tilde{j}}$ and collect the observation $Y_{x_{h_t,j_t}}$, $n=n+K$, $t=t+1$ }
	}
 \textbf{Return} the node according to the returning rule.\;
	\caption{StoROO}\label{sto}
\end{algorithm}
%In the experimental section, we illustrate the impact of different choices on the empirical efficiency.

% In this paper we do not provide indications for selecting this parameters. But in the experimental section we illustrate the impact of different choices.
	\subsection{Analysis of the algorithm}
In this section we provide a theoretical analysis of StoROO. It is inspired by~\cite{munos2014bandits}, but differs most notably by the fact that the analysis is suited for any $g$ and not only for the conditional expectation. The analysis relies on the possibility to construct, for any $\eta>0$, upper- and lower-confidence bounds $U_{h,j}^\eta(t)$ and $L_{h,j}^\eta(t)$ such that the event
$$\mathcal{A_\eta}=\bigcap_{T\geq t\geq 1}\bigcap_{\mathcal{P}_{h,j}\in \mathcal{T}_t}\Big\{U_{h,j}^\eta(t)\geq g(x_{h,j}),~L_{h,j}^\eta(t)\leq g(x_{h,j})\Big\}$$
has probability $\mathds{P}(\mathcal{A_\eta})$ at least $ 1-\eta$. 
We defer to Section \ref{sec:quantile} their specific expression for the cases of the quantile and CVaR. Especially Section \ref{sec:quantile} shows that in our setting the size of the confidence interval associated to each node is not always explicit, by opposition of the classical case.
We thus need to introduce the following definition to quantify how many times a node needs to be sampled before satisfying the expansion condition (Eq. \ref{explose}).

\begin{definition}\label{conv}
	Let
	$m_{\eta,h}(\theta,\kappa,\alpha)=\log(\theta T^2/\eta) \bigg(\dfrac{\kappa}{\hat{\beta}\delta(h)^{\hat{\gamma}}}\bigg)^\alpha $ \text{ and }
	$N_{h,j}(t)=\sum_{s=1}^{t}\mathds{1}_{X(s)\in \mathcal{P}_{h,j}}$,
	a \emph{vector of safe constants $v=(\theta, \kappa,\alpha)$} is composed of constants $\theta>0$, $\kappa>0$,  and $\alpha>0$ such that the event
	$$\mathcal{B}_\eta=\bigcap_{T\geq t\geq 1}\bigcap_{N_{h,j}\geq m_{\eta,h}(\theta,\kappa,\alpha)}\bigcap_{\mathcal{P}_{h,j}\in \mathcal{T}_t}\Big\{U_{h,j}^\eta(t)-L_{h,j}^\eta(t)\leq\hat{\beta}\delta(h)^{\hat{\gamma}}\Big\}$$
	has probability at least $1-\eta$. 
\end{definition}
For example, in the case of the conditional expectation a direct consequence of Hoeffding's inequality provides $\theta=2$, $\alpha=2$ and $\kappa=\sqrt{1/2}$ (see \cite{munos2014bandits}).% and
%$n_{\eta,h}=\dfrac{\log( T^2/\eta)}{2(\beta\delta(h)^\gamma)^{2}}.$

To ensure the convergence of StoROO, we first prove (Proposition \ref{expand}) that any point at the center of an expanded cell of depth $h$ belongs to
\begin{equation}
J_h=\{ ~x_{h,j} ~\text{such that}~ g(x_{h,j})+ 2 \hat{\beta}\delta(h)^{\hat{\gamma}}\geq g^*\}\;.
\end{equation}
Next, Proposition \ref{budg} shows that using a budget $T$, the tree $\mathcal{T}_T$ reaches at least a depth $H_\eta^*(T)$. 
This implies the point returned by the algorithm belongs to $J_{H_\eta^*(T)}$ (Proposition \ref{regret}). 
Finally, using an assumption on the size of $J_h$ that can be formalized by the so-call \textit{near-optimality dimension}, we provide an upper bound on the regret (Theorem \ref{main_doo}).

\begin{proposition}\label{expand}
	Conditionally on $\mathcal{A_\eta}$, StoROO only expands cells
	$\mathcal{P}_{h,j}$ such that $x_{h,j}\in J_h$.
\end{proposition}
Given the safe constants $v$ and the total budget $T$, the deeper the algorithm builds the tree, the better are the guarantees on the final point returned. 
So the goal of the following proposition is to provide a lower bound on the depth of $\mathcal{T}_T$.
\begin{proposition}\label{budg}
	Define $n_{\eta,h}=m_{\eta,h}(v)$ and define $H_{\eta}$ the largest $h\in \mathds{N}$ such that 
	$$S_{h}=K \sum_{h'\leq h}n_{\eta,h'+1}|J_{h'}|\leq T,~~\text{with}~~ |J_{h'}|\text{the cardinal of } J_{h'}\;.$$
	The deepest node $H_{\eta}^*$ expanded by StoROO is such that
	$H_{\eta}^*\geq H_{\eta}.$
	%Running qOO-1 with budget $T$ implies there exists at least one node expanded at depth $H^*$.
\end{proposition}
Intuitively, $S_{h}$ is the budget needed to expand all the nodes in $J_{h}$ for all $h'\leq h$. It may be that some of this nodes will not be visited, but in the worst case they are and they need to be considered in order to obtain a valid bound.
Putting Propositions \ref{expand} and \ref{budg} together, yields a first upper bound on the simple regret:
\begin{proposition}\label{regret}
	Running StoROO with budget $T$, with probability $\mathds{P}(\mathcal{A}_\eta\cap\mathcal{B}_\eta)$ the regret is bounded as
	$$r_T\leq 2 \hat{\beta} \delta\big(H_\eta^*(T)\big)^{\hat{\gamma}}\;.$$
\end{proposition}

A more explicit bound for the regret can be obtained by quantifying the volume of
$\mathcal{X}_\epsilon=\{x\in\mathcal{X},~g(x)\leq g^*-\epsilon \}$
for small values of $\epsilon$. Introducing the Holderian semi-metric
$\ell_{\beta,\gamma}(x,x')=\beta \left\| x-x'\right\| ^\gamma,$
that is associated with its regularity constants $\beta$ and $\gamma$, the \textit{near-optimality} dimension of the function is defined as follows, (see~\citet{munos2014bandits,bubeck2011x} for more details).
\begin{definition}\label{nearoptimality}
	The $\nu$-near optimality dimension is the smallest $d\geq 0$ such that for all $\epsilon\geq 0$, there exists $C\geq 0$ such that the maximal number of disjoint $\ell_{\hat{\beta},\hat{\gamma}}$-balls of radius $\nu \epsilon$ with center in $\mathcal{X}_\epsilon$ is less than $C\epsilon^{-d}$.
\end{definition}
% The near optimality dimension is a notion that depends on $g$ and on the semi metric $l_{\beta,\gamma}$, see
In order to evaluate $H_\eta^*$, we need to bound $|J_h|$ for all $h\geq 0$. The following proposition makes the link between the near optimality dimension and $|J_h|$.
\begin{proposition}\label{nearopti}
	Let $d$ be the $\frac{\nu^{\hat{\gamma}}}{2}$-near-optimality dimension, and $C$ the corresponding constant. Then 
	$$|J_h|\leq \frac{C}{\big(2 \hat{\beta} \delta(h)^{\hat{\gamma}}\big)^{d}}\;.$$
\end{proposition}
Finally, combining Propositions \ref{regret} and \ref{nearopti} with an hypothesis on the decreasing sequence $\delta(h)$, it is possible to provide the speed of convergence of $r_T$.
\begin{theorem}\label{main_doo}
	Assume that $\delta(h)=c\rho^{ h}$  for some $c\geq 0$ and $\rho< 1$, and assume that $v=~(\theta,\kappa,\alpha)$.
	Thus with probability $\mathds{P}(\mathcal{A}_\eta\cap \mathcal{B}_\eta)$, the regret of StoOO is bounded as
	$$r_T\leq c_1\Big[\dfrac{ \log(\theta T^2/\eta)}{T}\Big]^{\frac{1}{ d+\alpha}}
	\quad\text{ with } 	\quad
	c_1=2\hat{\beta}\bigg[\dfrac{KC\kappa^\alpha [2 \hat{\beta} ]^{-d}}{(1-\rho^{ d\hat{\gamma}+\hat{\gamma} \alpha})}\bigg]^{\frac{1}{d+ \alpha}},$$
	where $d$ is the near optimality dimension and $C$ the corresponding near optimality constant.
\end{theorem}
If $g$ is the conditional expectation, a vector of \textit{safe constants} is $(\theta=2, \alpha=2, \kappa=\sqrt{1/2})$ (based on Hoeffding's inequality). Thus if we plug it into the quantity defined in Theorem \ref{main_doo} we obtain
$$r_T\leq c_1\Big[\dfrac{ \log(2 T^2/\eta)}{T}\Big]^{\frac{1}{ d+2}}
	\quad\text{ with } \quad
	c_1=2\hat{\beta}\bigg[\dfrac{KC [2 \hat{\beta} ]^{-d}}{2(1-\rho^{ d\hat{\gamma}+\hat{\gamma} \alpha})}\bigg]^{\frac{1}{d+ 2}},$$
that is equivalent to what it is obtained in \cite{munos2014bandits}. \\
\textbf{Remark:} In the particular case where each cell is a hypercube and 
the sub-regions are created by the division of the parent-cell into $K=2^D$ sub-regions of equal size, then $K=2^D$, $c$ is equal to $\sqrt{D}$ and $\rho$ is equal to $\frac{1}{2}$.

	\section{Optimizing quantiles}\label{sec:quantile}
In this section, we focus on the optimization of \emph{quantiles}, which are well-established tools in (risk-averse) decision theory \citep[see][for instance]{rostek2010quantile}. 
In particular, they benefit from interesting robustness properties, with respect to outliers or heavy tails.
% 	Motivated by some qualities of the quantile, for exemple its robustness to outliers  for more details about the utilization of quantile in decision theory), here we focus our interest on the quantile as a mesure of risk.
Let  $$g(x)=q_x(\tau)=\inf\big\{q\in\mathds{R}: F_x(q)\geq \tau \big\}\;,$$ be the $\tau$-quantile of $Y_x$, where $F_x$ is the cumulative distribution function (CDF) of~$\mathds{P}_x$.
Here we detail how to construct the UCB and LCB for quantiles. 
First, we provide bounds based on Hoeffding's inequality %\citep{hoeffding1994probability} 
and we use them to adapt the regret bounds of Theorem \ref{main_doo}. 
Then we provide two more refined bounds that take into account the order $\tau$ of the quantile based respectively on the Bernstein's inequality and on the Kullback-Leibler divergence. 
% Finally we highlight the interest of these bounds (especially when $\tau$ is far from $0.5$) both theoretically and numerically. 

Let us first introduce some notations. 
For all $1\leq t\leq T$, $1\leq h\leq t$, $1\leq j\leq K^h$ and $q \in \mathds{R}$ we denote
% 	\aurelien{$N_{h,j}(t)$ est dÃÂÃÂ©jÃÂÃÂ  dÃÂÃÂ©fini plus haut non ? cf Definition 2}
\begin{equation*}
% 	N_{h,j}(t)=\sum_{s=1}^{t}\mathds{1}_{X(s)\in \mathcal{P}_{h,j}} \quad \text{ and } 
\hat{F}_{h,j}^{t}(q) =\dfrac{\sum_{s=1}^{t} \mathds{1}_{Y(s)\leq q}\mathds{1}_{X(s)\in \mathcal{P}_{h,j}}}{N_{h,j}(t)},
\end{equation*}
the empirical CDF of the reward inside the cell $\mathcal{P}_{h,j}$, where $N_{h,j}(t)$ is the (random) number of times the cell was sampled up to time $t$ (see Definition~\ref{conv}). 
% 	
% 	$Z_{h,j}^{\tau}(t)=\mathds{1}_{Y(t)\leq q_{x_{h,j}}(\tau)}\mathds{1}_{X(t)\in \mathcal{P}_{h,j}}$ and
% 	$$S_{h,j}^\tau(t)=\sum_{s=1}^{t} Z_{h,j}^{\tau}(s),~~~~N_{h,j}(t)=\sum_{s=1}^{t}\mathds{1}_{X(s)\in \mathcal{P}_{h,j}},~~~~ \hat{F}_{h,j}^{t}(q_\tau) =\dfrac{S_{h,j}^\tau(t)}{N_{h,j}(t)},$$
% 	
% 	 with $q_{x_{h,j}}(\tau)$ the quantile of order $\tau$ of $\mathds{P}_{x_{h,j}}$.
The \textit{generalized inverse} $\hat{F}_{h,j}^{t~-}$ of the piecewise constant function $\hat{F}_{h,j}^{t}$ is defined as $\hat{q}_{h,j}(\tau)=\inf\big\{q\in\mathds{R}:~\hat{F}_{h,j}^{t}(q)\geq \tau\big\},$  %=y_{([N_{h,j}\tau])}(x_{h,j}),
that is the $\lceil N_{h,j}(t)\times \tau \rceil$ order statistic of the sample that has been collected from the node $x_{h,j}$ until time $t$.

\begin{figure}[ht!]
	\centering
	\begin{tabular}{cc}
		\includegraphics[width=9cm]{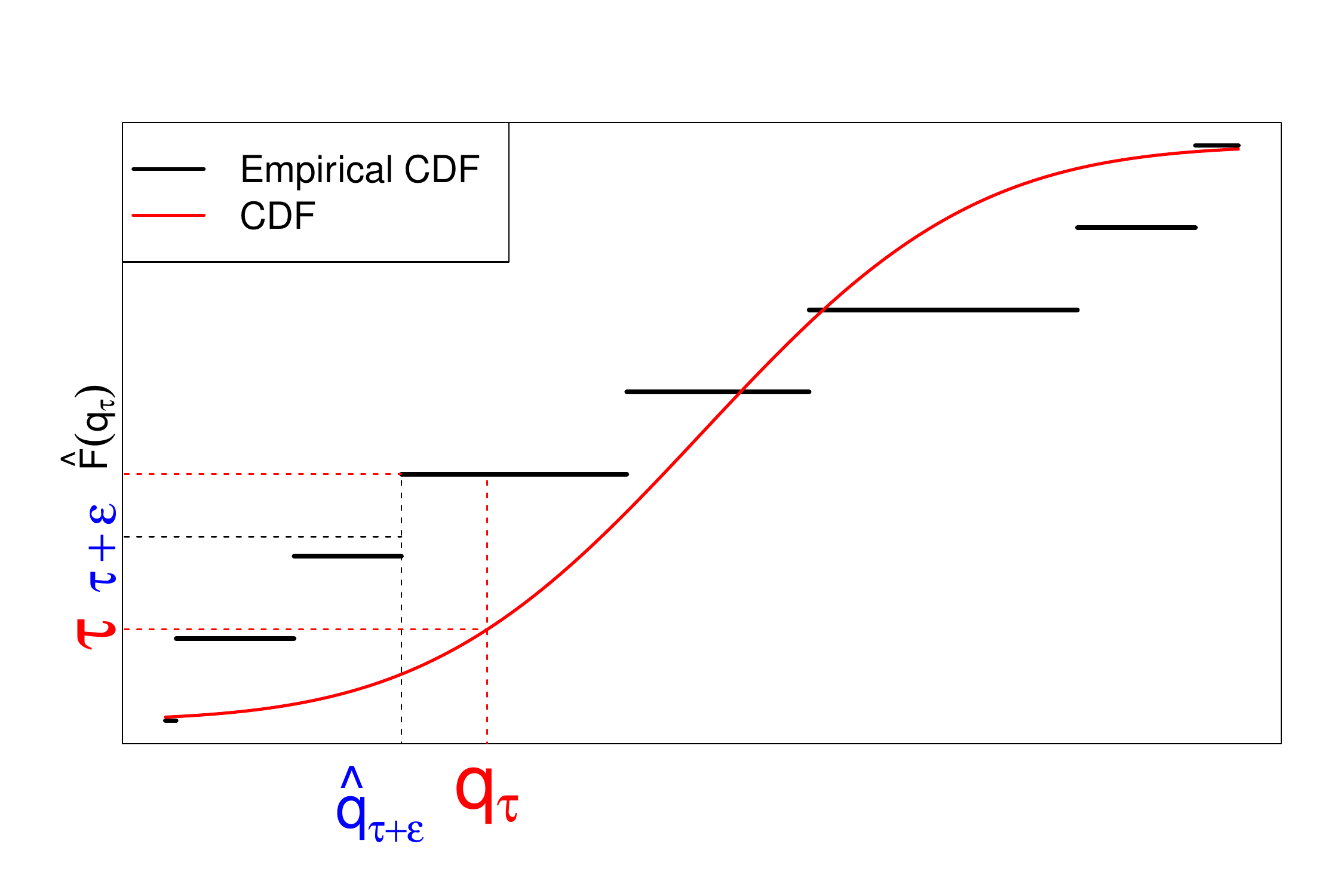}
	\end{tabular}
	\caption{Illustration of the equivalence (\ref{ucb1}).}\label{fig:equivalence}
\end{figure}

To define confidence bounds on the conditional quantile we proceed in two steps. First we propose confidence bounds on $\hat{F}_{h,j}(q_\tau)$.
To do so, we simply use deviation bounds for Bernoulli distributions, since for all $x\in\mathcal{X}$, for all $1\leq n \leq T$, the random variables $\big(\mathds{1}_{Y_x (\xi_s)\leq q_{x}(\tau)}\big)_{s=1,\cdots,n}$ 
are independent and identically distributed with a Bernoulli law of parameter $\tau$, if $\xi_s$ denotes the time when the node $x$ has been sampled for the $s$-th time.
Then we use the properties
\begin{eqnarray}
\forall~~\epsilon>0~~\text{such that}~~\tau+\epsilon<1,\quad \hat{F}_{h,j}^{t}\big(q_{h,j}(\tau)\big)\geq\tau+\epsilon &\Leftrightarrow&q_{h,j}(\tau)\geq \hat{F}_{h,j}^{t~-}(\tau+\epsilon)\;,\label{ucb1}\\
\forall~~\epsilon>0~~\text{such that}~~\tau+\epsilon>0,\quad \hat{F}_{h,j}^{t}\big(q_{h,j}(\tau)\big)<\tau-\epsilon &\Leftrightarrow& q_{h,j}(\tau)\leq \hat{F}_{h,j}^{t~-}(\tau-\epsilon)\;,\label{ucb2}
\end{eqnarray}
to create confidence bounds on $q_{h,j}(\tau)$ using bounds on $\hat{F}_{h,j}^t(q_\tau)$. Note that here we just assume   that the output distribution has a continuous, striclty increasing cumulative distribution function. It is not necessary to assume something else, such as bounded support or bounded moments because here we refer to Bernouilli distributions. The first equivalence in illustrated on Figure \ref{fig:equivalence}.

%\begin{figure}
%	\centering
%	\begin{tabular}{cc}
%		\includegraphics[width=9cm]{cdf_ucb.pdf}
%	\end{tabular}
%	\caption{Illustration of the equivalence (\ref{eq-UCB}).}\label{fig:equivalence}
%\end{figure}

\subsection{Hoeffding's bound and regret analysis}

Let $\epsilon_{N_{h,j}(t)}^{\eta,T}=\sqrt{\dfrac{\log(2T^2/\eta)}{2N_{h,j}(t)}}$, and let
\begin{equation}\label{UCB}
U_{h,j}^{\eta}(t)= \left\{
\begin{array}{ll}
\min\big\{q,~\hat{F}_{h,j}^{t}(q)\geq \tau+\epsilon_{N_{h,j}(t)}^{\eta,T}\big\} & \mbox{\text{if} $\tau+\epsilon_{N_{h,j}(t)}^{\eta,T}<  1$} \\
+\infty & \mbox{otherwise,}
\end{array}
\right.
\end{equation}
\begin{equation}\label{LCB}
L_{h,j}^{\eta}(t)= \left\{
\begin{array}{ll}
\max\big\{q,~\hat{F}_{h,j}^{t}(q)\leq \tau-\epsilon_{N_{h,j}(t)}^{\eta,T}\big\} & \mbox{\text{if} $\tau-\epsilon_{N_{h,j}(t)}^{\eta,T} > 0$} \\
-\infty & \mbox{otherwise.}
\end{array}
\right.
\end{equation}
The next proposition motivates the choice of the above quantities as a UCB and a LCB for the quantile of order $\tau$ at the points $(x_{h,j})_{(h,j)\in\mathcal{T}_t}$.
\begin{proposition}\label{hoeffding}
Assume that for all $x\in\mathcal{X}$, $\mathds{P}_x$ has a continuous, striclty increasing cumulative distribution function then for any $\eta>0$, for all  $h\geq 0$, for all $0\leq j \leq K^h$ and for all $1\leq t\leq T$, 
if $L_{h,j}^{\eta}(t)$  and $U_{h,j}^{\eta}(t)$ are defined according to (\ref{LCB}) and (\ref{UCB}), respectively, then the event $\mathcal{A}_\eta$ has probability at least $1-\eta$.
	% 		\aurelien{"the event A holds with probability $1-\delta$" ÃÂÃÂ§a ne veut rien dire}
\end{proposition}
Now, analyzing the regret requires a high probability bound on the number of time a node is sampled before being expanded:
%\begin{proposition}\label{Massart}
%Under the conditions required by Proposition \ref{hoeffding}, define $f_x$ as the density of %$\mathds{P}_x$ and define $\bar{f}(x)=\min_{y\in\supp Y_x }f_x(y)$. If $L_{h,j}^{\eta}(t)$  and %$U_{h,j}^{\eta}(t)$ are defined according to (\ref{LCB}) and (\ref{UCB}), respectively then a vector %of safe constants is given as
	%$v=\left(2,\frac{2\sqrt{2}}{\min_{x\in\mathcal{X}}\bar{f}(x)},2\right),$
	%and for any $\eta>0$, $\mathds{P}(\mathcal{A}_\eta\cap\mathcal{B}_\eta)\geq 1-\eta.$
%\end{proposition}
\begin{proposition}\label{Massart}
Under the conditions required by Proposition \ref{hoeffding}, define $f_x$ as the density of $\mathds{P}_x$ and define $\bar{f}(x)=\min_{\tau'\in [\tau-2\epsilon_{M_{\tau}}^{\eta,T},\tau+2\epsilon_{M_\tau}^{\eta,T}]} f_{x} \circ F_{x}^{-1}(\tau')$ with $M_\tau=2m_\tau^{-2}\log(2T^2/\eta)$ and $m_\tau= \min(\tau,1-\tau)$.
%$$m_\tau=\max\bigg(\dfrac{2\log(2T^2/\eta)}{\tau^2}, \dfrac{2\log(2T^2/\eta)}{(1-\tau)^2}\bigg).$$
If $U_{h,j}^{\eta}(t)$ and $L_{h,j}^{\eta}(t)$ are defined according to (\ref{UCB}) and (\ref{LCB}), respectively, then for any $\eta>0$, $\mathds{P}(\mathcal{A}_\eta\cap\mathcal{B}_\eta)\geq 1-\eta$ and a vector of safe constants is given as
$$v=\left(2,\frac{\sqrt{8 m_\tau^2+ 4\big(\hat{\beta}\diam(\mathcal{X})^{\hat{\gamma}}\min_{x\in\mathcal{X}}\bar{f}(x)\big)^2}}{m_\tau\min_{x\in\mathcal{X}}\bar{f}(x) },2\right).$$
\end{proposition}

According to the previous proposition, if we have sampled a node at depth $h$ more than
	\begin{equation} 
	%n_{\eta,h}(\kappa,\alpha)
	n_{\eta,h}=\log(2 T^2/\eta) \bigg(\frac{8m_\tau^2+ 4\big(\hat{\beta}\diam(\mathcal{X})^{\hat{\gamma}}\min_{x\in\mathcal{X}}\bar{f}(x)\big)^2}{\big(\min_{x\in\mathcal{X}}\bar{f}(x)m_\tau \hat{\beta}\delta(h)^{\hat{\gamma}}\big)^2}\bigg)\label{budgetquantile}
	\end{equation}
times, then with probability $1-\eta$, Condition~\eqref{explose} is satisfied and thus the node is expanded. 
	
Equality~(\ref{budgetquantile}) reflects two dependencies. The smaller the minimum of the density over a neighborhood of the quantile and the closer $\tau$ from $0$ or $1$, the larger the upper bound on the number of samples needed before being expanded. Indeed a small density value in a neighborhood of the targeted quantile will produce samples with few observations close to the quantile, hence the estimation error will be large. In addition from Proposition~(\ref{hoeffding}), to obtain non trivial UCB and LCB, the value $N_{h,j}$ has to be large enough to ensure $\tau\pm\epsilon_{N_{h,j}}^{\eta,T}\in[0,1]$ and this value increases as $\tau$ comes close from $0$ or $1$.  %The first dependency is not very surprising as the central limit theorem for sample quantile propose a rate of convergence depending on the inverse of the density at the quantile value. But it is clear as well that the closer tau from $0$ and $1$, le larger
%Actually the bound is crude. It is rather clear, in fact, that the \emph{local} minimum of $f_x$ around $q_x(\tau)$ is the crucial quantity. Here we chose to write the results in terms of the global minimum to simplify the proof of Proposition (\ref{Massart}). %, the lower bound is taken as $\min_{x\in\mathcal{X}}\bar{f}(x)$, however that choice hides the dependence in $q_x(\tau)$. 
Thus a more precise way to understand the behaviour of StoROO  is that the number of time  a node needs to be sampled before expansion depends on the pdf value in a neighborhood (of decreasing size with $N_{h,j}$) of the targeted quantile.
	
To obtain an upper bound on the simple regret, we now just need to combine Theorem~\ref{main_doo} with Proposition~\ref{Massart} so as to obtain the following theorem.

\begin{theorem}\label{quantile_regret}
	Under the conditions required by Proposition \ref{hoeffding} and \ref{Massart}, if $\delta(h)=c\rho^{ h}$  for some $c\geq 0$ and $\rho< 1$, then with probability $1-\eta$, the regret of StoROO for maximizing the quantile is bounded as
	$$r_T\leq c_2\Big[\dfrac{ \log(2T^2/\eta)}{T}\Big]^{\frac{1}{ d+2}}~\text{with}~c_2^{d+2}=KC\hat{\beta}^2\dfrac{ 16m_\tau^2+ 8\big(\hat{\beta}\diam(\mathcal{X})^{\hat{\gamma}}\min_{x\in\mathcal{X}}\bar{f}(x)\big)^2 }{\big(m_\tau\min_{x\in\mathcal{X}}\bar{f}(x) \big)^2(1-\rho^{ d\hat{\gamma}+ \hat{\gamma} \alpha})}\;,$$
	with $d$ the near-optimality dimension and $C$ the near-optimality corresponding constant.
	%$$r_T\leq O\Bigg(\Big[\dfrac{ \log(T^2/\eta)}{T}\Big]^{\frac{1}{ 2+d}}\Bigg).$$
\end{theorem}
Note that the speed of convergence is the same as the one obtained in the conditional expectation optimization setting; only the constant varies.
	\subsection{Tighter bounds}
Using Hoeffding's inequality is convenient because it leads to explicit lower and upper confidence bounds, which simplifies the deriviation of bounds on the regret.
	However, it implicitly upper-bounds the variance of all $[0,1]$-valued random variables by $1/4$, which is overly pessimistic when the inequality is applied to variables whose expectations are far from $1/2$.
	This is in particular the case for quantile estimation, when the quantile is of order close to $0$ or $1$. 
	To take into account the order of the quantile, following \cite{david2016pure}, a first possibility is to derive confidence intervals from Bernstein's inequality as presented in the following proposition.
				\begin{proposition}\label{unionb}
					For any $\eta>0$, for all $1\leq t\leq T$, $1\leq h\leq t$ and $1\leq j\leq K^h$, define
					\begin{equation}
					U_{h,j}^\eta(t)= \left\{
					\begin{array}{ll}
					\min\big\{q,~\hat{F}_{h,j}^{t}(q)\geq \tau+\epsilon_{N_{h,j}(t)}^{\eta,T}\big\} & \mbox{\text{if} $\tau+\epsilon_{N_{h,j}(t)}^{\eta,T}<  1$} \\
					+\infty & \mbox{otherwise,}\nonumber
					\end{array}
					\right.
					\end{equation}
					and
					\begin{equation}
					L_{h,j}^\eta(t)= \left\{
					\begin{array}{ll}
					\max\big\{q,~\hat{F}_{h,j}^{t}(q)\geq \tau-\epsilon_{N_{h,j}(t)}^{\eta,T}\big\} & \mbox{\text{if} $\tau-\epsilon_{N_{h,j}(t)}^{\eta,T}> 0$} \\
					-\infty & \mbox{otherwise,}\nonumber
					\end{array}
					\right.
					\end{equation}
					with
					$$\epsilon_{N_{h,j}(t)}^{\eta,T}=\dfrac{\log(2T^2/\eta)}{3N_{h,j}(t)}\bigg(1+\sqrt{1+\dfrac{18N_{h,j}(t)\tau(1-\tau)}{\log(2T^2/\eta)}}\bigg).$$
					%Define $\mathcal{A}_\eta$
					%$$\mathcal{A}_\eta=\bigcap_{T\geq t\geq 1}\bigcap_{\mathcal{P}_{h,j}\in \mathcal{T}_t}\Big\{U_{h,j}^\eta(t)\geq q_{h,j}(\tau),~L_{h,j}^\eta(t)\leq q_{h,j}(\tau)\Big\}.$$
					If $g$ is the conditional quantile of order $\tau$ then the event $\mathcal{A}_\eta$ has probability at least $1-\eta$.
				\end{proposition}

 Although Bernstein's inequality takes into account the order of the quantile, it is possible to do something better.
In order to create tighter confidence bounds, we thus go back to Chernoff's inequality and derive less explicit, but more accurate upper- and lower- confidence bounds on the $\tau$-quantiles. We follow here~ \cite{garivier2011kl}, but a close inspection at the proofs shows however a difference in the order of the marginals of the KL functions.
	Recall that the binary relative entropy is defined for $(p,q)\in [0,1]^2$ as:
	$$\kl(p,q)=p\log\dfrac{p}{q}+(1-p)\log\dfrac{1-p}{1-q}\;,$$
	with by convention, $0\log0=0$, $\log0/0=0$ and $x\log x/0=+\infty ~\text{for}~ x>0.$
	\begin{proposition}\label{union}
		For any $\eta>0$, for all $1\leq t\leq T$, $1\leq h\leq t$ and $1\leq j\leq K^h$, define
	$$U_{h,j}^\eta(t)=\min\Big\{q,~\hat{F}_{h,j}^n(q)\geq\tau~\text{and}~\kl(\hat{F}_{h,j}^t(q),\tau)\geq \dfrac{\log(2T^2/\eta)}{N_{h,j}(t)}\Big\}~\text{if}~\kl(1,\tau)>\dfrac{\log(2T^2/\eta)}{N_{h,j}(t)}$$
$\text{and}~+\infty~\text{otherwise.}$ Define \\
$$L_{h,j}^\eta(t)=\max\Big\{q,~\hat{F}_{h,j}^t(q)\leq\tau~\text{and}~\kl(\hat{F}_{h,j}^t(q),\tau)\geq \dfrac{\log(2T^2/\eta)}{N_{h,j}(t)}\Big\}~\text{if}~\kl(0,\tau)> \dfrac{\log(2T^2/\eta)}{N_{h,j}(t)}$$
$\text{and}~-\infty~\text{otherwise.}$
		%Define $\mathcal{A}_\eta$
		%$$\mathcal{A}_\eta=\bigcap_{T\geq t\geq 1}\bigcap_{\mathcal{P}_{h,j}\in \mathcal{T}_t}\Big\{U_{h,j}^\eta(t)\geq q_{h,j}(\tau),~L_{h,j}^\eta(t)\leq q_{h,j}(\tau)\Big\}.$$
		Then the event $\mathcal{A}_\eta$ has probability at least $1-\eta$.
	\end{proposition}
Contrary to Bernstein's inequality, Chernoff's bound
	 is always tighter than Hoeffding's inequality, which follows from Pinsker's inequality \citep[see e.g.][]{garivier2018explore}. It follows in particular that the regret of StoROO using confidence bounds derived from Chernoff's inequality has, at least, the guarantees presented in Theorem \ref{quantile_regret}.
	
	\iffalse
 Contrary to Bernstein's inequality, Chernoff's bound $\mathds{P}\big(\hat{F}^n(q(\tau))\geq x\big)\leq \exp(-n \kl(x,\tau))$ 
	 is always tighter than Hoeffding's inequality $\mathds{P}(\hat{F}^n\big(q(\tau))\geq x\big)\leq \exp\big(-2n (\tau-x)^2\big)$, which follows from Pinsker's inequality \citep[see e.g.][]{garivier2018explore}:
	$$\forall~0\leq p<q \leq 1,~ \kl(p,q)\geq  \dfrac{1}{2\max_{x\in[p,q]}x(1-x)}(p-q)^2\geq 2(p-q)^2\;.$$
	For example, given $\tau>0.5$ and an i.i.d. sample of size $n$, one can see that % using the Pinsker's inequality and the equivalence (\ref{eq-UCB}), it is easy to see that 
	$$U^{\kl}_n\leq \hat{q}_n\left(\tau+\sqrt{\dfrac{2\tau(1-\tau)\log(2/\eta)}{n}}\right)<\hat{q}_n\left(\tau+\sqrt{\dfrac{\log(2/\eta)}{2n}}\right)=U^{\h}_n,$$
	with $U^{\kl}$ (resp. $U^{\h}$) the UCB associated to Chernoff's inequality (resp. Hoeffding's inequality). Bernstein's inequality is tighter than Hoeffding's when $\tau$ is different from $1/2$ and $n$ sufficiently large, but always looser than Chernoff.
	\fi

	%In addition using the same indices
	%$$L_{\kl}(n)\geq q \left(\sqrt{\dfrac{\tau-\log(2/\eta)}{2n}} \right)=L_{\h}(n).$$
	%For $\tau<0.5$, by replacing the LCB with the UCB, it is possible to obtain the same guarantees. For $\tau=0.5$ both bounds are close.
	%\victor{on se demande pourquoi il n'y a pas de discussion ici avec Bernstein}\leonard{J'ai ajoutÃ© en tout dÃ©but de paragraphe qu'il n'y avait de relations strictes entre Bernstein et Hoeffding. Ca suffit ou c'est maladroit ?}

The online setting we consider in this article induces that, after $t$ steps, the set of nodes and the number of observations in each node are random. To cope with this, we thus need deviation bounds for random size samples. The most simple way to obtain such inequalities is to use a union bound on the possible number of observations in each node, as presented above. Tighter results can be obtained from a more thorough analysis (sometimes called \textit{peeling trick}): this is what is presented below.
	\begin{proposition}\label{peeling}
	For any $\eta\in(0,1)$ let
		$\displaystyle{\delta_\eta(T)=\inf\big\{\delta>0 : ~T e \lceil \delta \log(T) \rceil \exp(-\delta)\leq \eta/2\big\}}$,
		and define
			$$U_{h,j}^\eta(t)=\min\Big\{q,~\hat{F}_{h,j}^n(q)\geq\tau~\text{and}~N_{h,j}(t)\kl(\hat{F}_{h,j}^t(q),\tau)\geq \delta_\eta(T)\Big\}~\text{if}~\kl(1,\tau)>\dfrac{\delta_\eta(T)}{N_{h,j}(t)}$$
$\text{and}~+\infty~\text{otherwise.}$ Define
	$$L_{h,j}^\eta(t)=\max\Big\{q,~\hat{F}_{h,j}^t(q)\leq\tau~\text{and}~N_{h,j}(t)\kl(\hat{F}_{h,j}^n(q),\tau)\geq \delta_\eta(T)\Big\}~\text{if}~\kl(0,\tau)> \dfrac{\delta_\eta(T)}{N_{h,j}(t)}$$ 
$\text{and}~-\infty~\text{otherwise.}$
 		%The event %$\mathcal{A}_t$
 		%$$\mathcal{A}_\eta=\bigcap_{\mathcal{P}_{h,j}\in \mathcal{T}_t}\Big\{U_\eta^{h,j}(t)\geq q_{h,j}(\tau),~L_\eta^{h,j}(t)\leq q_{h,j}(\tau)\Big\}$$
 	Then the event $\mathcal{A}_\eta$ has probability at least $1-\eta$.
 \end{proposition}
Note that for every $0<\delta\leq \log(2/\eta)$, 
$\lceil \delta \log(T) \rceil\geq 1$ and thus $ T e \lceil \delta \log(T) \rceil \exp(-\delta) > \eta/2$; hence,  $\delta_\eta(T)>\log(2/\eta)$.

% 	
% 	In the case $\tau< 0.5$, we can show similarly that the LCB is improved and that the UCB is at least as good as what we have with the Hoeffding's inequality. 
% 	If $\tau=0.5$ we can only say that the new bounds are not worse than Hoeffding's. This implies that the regret is at least the same using KL. 
% 	\aurelien{oui bon lÃ  non plus je ne sais pas si c'est la peine de le dire, sinon juste "of course, a similar result holds for $\tau<0.5$, and both bounds are close when $\tau=0.5$.}
% 	\begin{figure}
% 		\begin{tabular}{cc}
% 			\includegraphics[width=7cm]{kl05.pdf}
% 			\includegraphics[width=7cm]{kl09.pdf}
% 		\end{tabular}
% 		\caption{Difference between the kl and the exponent in the case of Hoeffding \victor{phrase pas rigoureuse et legende incomplete : toutes les lignes doivent etre decrites. A ameliorer seulement si on garde la figure}\aurelien{pour moi le dessin est mÃªme Ã  enlever si on est juste en place: c'est dÃ©jÃ  bien trop long sur une idÃ©e qui n'est pas nouvelle}}
% 	\end{figure}
		\section{Optimizing CVaR}
We now detail how StoROO can be applied to the optimization of another important notion of risk: the CVaR. CVaR has raised a great interest in recent years, notably because it is a \textit{coherent} risk indicator (see \cite{ben2007old} for instance). For $\tau\in[0,1)$ the condition value at risk at level $\tau$ of a continuous random variable $Y$ is defined as
	$$\cvar(Y)=\inf_{z\in\mathds{R}}\Big\{z+\dfrac{1}{(1-\tau)}\mathds{E}[(Y-z)^+]\Big\}=\mathds{E}\Big(Y|Y\geq q(\tau)\Big)\;,$$
	with $(z)^+=\max(0,z)$.
Following \cite{brown2007large}, it can be estimated by
$$\widehat{\cvar}^n=\inf_{z\in\mathds{R}}\Big\{z+\dfrac{1}{(1-\tau) n}\sum_{i=1}^{n}(Y_i-z)^+\Big\}=Y_{(\lfloor n\tau \rfloor)}+\dfrac{1}{(1-\tau)n}\sum_{i=1}^n(Y_i-Y_{(\lfloor n\tau \rfloor)})^+\;.$$

%Note that the second equality can be demonstrated using the fact that $\widehat{\cvar}^n$ is piecewise convex and that the slope is negative for $z<Y_{(\lfloor n\tau \rfloor)}$ and positive for $z>Y_{(\lfloor n\tau \rfloor)}$.
Since $Y$ often stands for a loss, the CVaR is usually to be minimized. 
In order to stay consistent with the rest of the paper, we choose in the following to maximizing $g= -\cvar$.

	Assuming the random variables are bounded in an interval $[a,b]$, the next proposition adapts the deviation inequalities presented in \cite{brown2007large} to our sequential setting.
	
\begin{proposition}\label{Brown}
For any $\eta>0$, for all  $h\geq 0$, for all $0\leq j \leq K^h$ and for all $1\leq t\leq T$, 
define \begin{equation}\label{ucbcvar}
	    U_{h,j}^{\eta}(t)= -\widehat{\cvar}^t(h,j)+ \dfrac{b-a}{1-\tau}\sqrt{\dfrac{\log(2T^2/\eta)}{2 N_{h,j}(t)}},\nonumber
	\end{equation}
	\begin{equation}
	L_{h,j}^{\eta}(t)= -\widehat{\cvar}^t(h,j)-(b-a)\sqrt{\dfrac{5\log(6T^2/\eta)}{(1-\tau) N_{h,j}(t)}} 	    ,\nonumber
	\end{equation}
	with 
	$$\widehat{\cvar}^t(h,j)=Y_{(\lfloor N_{h,j}(t)\tau \rfloor)}^{h,j}+\dfrac{1}{(1-\tau) N_{h,j}(t)}\sum_{i=1}^t \mathds{1}_{X(i)\in \mathcal{P}_{h,j}}(Y_i-Y_{(\lfloor N_{h,j}(t)\tau \rfloor)}^{h,j})^+,$$
	where $Y_{(k)}^{h,j}$ represents the value of $Y_{(k)}$ for the node $(h,j)$.
 If the random variables $Y_x$ are bounded in $[a,b]$ for all $x\in\mathcal{X}$ and have continuous distribution functions, then the event $\mathcal{A}_\eta$ has probability at least $1-\eta$. 

\end{proposition}
Note that \textit{deviation inequalities} can be established for CVaR in sub-Gaussian or light-tailed cases (see \cite{kolla2019risk} for instance) but an assumption has to be made on the value of the pdf in a neighborhood of the $\tau$-quantile.

From Proposition (\ref{Brown}), one can see that whenever a node has been played more than $\displaystyle{m_{\eta,h}= \log(6T^2/\eta)(b-a)^2\bigg( \dfrac{1+\sqrt{10(1-\tau)}}{\sqrt{2}(1-\tau) \hat{\beta} \delta(h)^{\hat{\gamma}}}\bigg)^2}$ times, it has been expanded. Thus a possible associated vector of \textit{safe constants} is $v=\bigg(6,(b-a)\Big( \dfrac{1+\sqrt{10(1-\tau)}}{\sqrt{2}(1-\tau) \hat{\beta} \delta^{\hat{\gamma}}}\Big),2\bigg).$ Combining $v$ with Theorem \ref{main_doo} provides the following upper bound on the regret.

\begin{theorem}\label{regret_cvar}
	Under the conditions required by Proposition \ref{Brown}, if $\delta(h)=c\rho^{ h}$  for some $c\geq 0$ and $\rho< 1$, then with probability $1-\eta$, the regret of StoROO for minimizing $\cvar$ is bounded as
	$$r_T\leq c_3\Big[\dfrac{ \log(6T^2/\eta)}{T}\Big]^{\frac{1}{ d+2}}~~\text{with}~~c_3=2\hat{\beta} \bigg[\dfrac{\big(1+\sqrt{10(1-\tau)}\big)^2KC (b-a)^2 [2 \hat{\beta} ]^{-d}}{2(1-\tau)^2(1-\rho^{ d\hat{\gamma}+ \hat{\gamma} \alpha})}\bigg]^{\frac{1}{d+ 2}},$$
	with $d$ the near-optimality dimension and $C$ the near-optimality corresponding constant.
	%$$r_T\leq O\Bigg(\Big[\dfrac{ \log(T^2/\eta)}{T}\Big]^{\frac{1}{ 2+d}}\Bigg).$$
\end{theorem}
The inequalities obtained in Proposition \ref{Brown} are convenient because they lead to explicit lower and upper confidence bounds, which simplifies the derivation of bounds on the regret. However, as they are based on Hoeffding's inequality, they can be over-conservative. To obtain better bounds, \cite{thomas2019concentration} propose data-dependent inequalities derived from the \textit{Dvoretzky-Kiefer-Wolfowitz} inequality. The following proposition provides the UCB and LCB based on these inequalities.
	\begin{proposition}\label{lern-mil}
		Assume for all $x\in\mathcal{X}$, $Y_x$ is bounded by $(a,b)\in\mathds{R}^2$.
		For any $\eta\in(0,0.5]$, for all $1\leq t\leq T$, $1\leq h\leq t$ and $1\leq j\leq K^h$, define
		\begin{equation}
		L_{h,j}^\eta(t)= \dfrac{1}{1-\tau}\sum_{i=1}^{N_{h,j}(t)}(Y_{i+1}^{h,j}-Y_i^{h,j})\Big(\dfrac{i}{N_{h,j}(t)}-\sqrt{\dfrac{\log(2T^2/\eta)}{2N_{h,j}(t)}}-\tau\Big)^+-Y_{T+1}^{h,j}\nonumber
		\end{equation}
		and
		\begin{equation}\label{LCBtight}
		U_{h,j}^\eta(t)=  \dfrac{1}{1-\tau} \sum_{i=0}^{N_{h,j}(t)-1}(Y_{i+1}^{h,j}-Y_i^{h,j})\Big(\min\big\{1,\dfrac{i}{N_{h,j}(t)}+\sqrt{\dfrac{\log(2T^2/\eta)}{2N_{h,j}(t)}}\big\}-\tau\Big)^+-Y_{N_{h,j}(t)}^{h,j}, \nonumber
		\end{equation}
		with  $Y_0^{h,j}=a$ and $Y_{T+1}^{h,j}=b$. Then if $g=-\cvar$, the event $\mathcal{A}_\eta$ has probability at least $1-\eta$.
	\end{proposition}
	Although we do not propose an analysis of the regret based on this bounds, it is immediate to state that the upper bound on the regret is always smaller than the bound obtained in Theorem \ref{regret_cvar} because these inequalities are strictly tighter than Brown's inequalities. In the following section, we numerically highlight  the relevance of using these tight bounds.

	\section{Experiments}\label{expe}
	We empirically highlight the capacity of StoROO to optimize the conditional quantile and CVaR of a black-box function. Four versions of StoROO are compared for both cases.
	
	For the conditional quantile we compare StoROO using confidence bounds repectively derived from
	Hoeffding's, Bernstein's, Chernoff's inequalities (resp. denoted $\StoROO_{\h}$, $\StoROO_{\ber}$ and $\StoROO_{\kl}$) and Chernoff's inequality and the \textit{peeling trick} ($\StoROO_{\klp}$). 
% 	For the conditional quantile we compare $\StoROO_{\h}$ ($i.e.$ StoROO using confidence bounds derived from Hoeffding's inequality), $\StoROO_{\ber}$ (Bernstein's inequality),
% 	$\StoROO_{\kl}$ (Chernoff's inequality) and $\StoROO_{\klp}$ (Chernoff's inequality and the \textit{peeling trick}). 
	
	For the optimization of the conditional CVaR, 
	we compare the use of confidence bounds derived from Brown's inequality and from \cite{thomas2019concentration}. To use these inequalities we have to provide $(a,b)\in\mathds{R}^2$ that bound the output. Hence, we compare two cases: one where we provide conservative bounds for $(a,b)$ (here $(a,b)=(0,1)$), and one where we provide their actual values ($a_x=\min \supp(Y_x)$ and $b_x=\max \supp(Y_x)$, $i.e.$ the minimum and the maximum of the support of the conditional distribution). We denote the four variants 
	$\StoROO_{\br}$ (from Brown's inequality), $\StoROO_{\T}$ (from \cite{thomas2019concentration}), and $\StoROO_{\bro}$ and $\StoROO_{\To}$
	for their variants with oracle bounds.
% 	
% 	Here we first assume $(a,b)=(0,1)$ and we denote $\StoROO_{\br}$ ($i.e$ StoROO using confidence bounds derived from Brown's inequality) and  $\StoROO_{\T}$ ($i.e$ StoROO using confidence bounds derived from \cite{thomas2019concentration}). Then for all $x\in\mathcal{X}$ we test choice of bounds $a_x=\min \supp(Y_x)$ and $b_x=\max \supp(Y_x)$ ($i.e$ the minimum and the maximum of the support of the conditional distribution) and we denote $\StoROO_{\bro}$ ($i.e$ StoROO using confidence bounds derived from Brown's inequality and the oracle bounds) and $\StoROO_{\To}$ ($i.e$ StoROO using confidence bounds derived from \cite{thomas2019concentration} and those oracle bounds).
% 	and $\StoROO_{\klp}$ ($i.e$ StoROO using confidence bounds derived from Chernoff's inequality and the peeling trick).

\begin{figure}[ht!]
		\begin{tabular}{cc}
			\includegraphics[trim={0cm 1cm 0cm 2.0cm},clip,width=7.2cm]{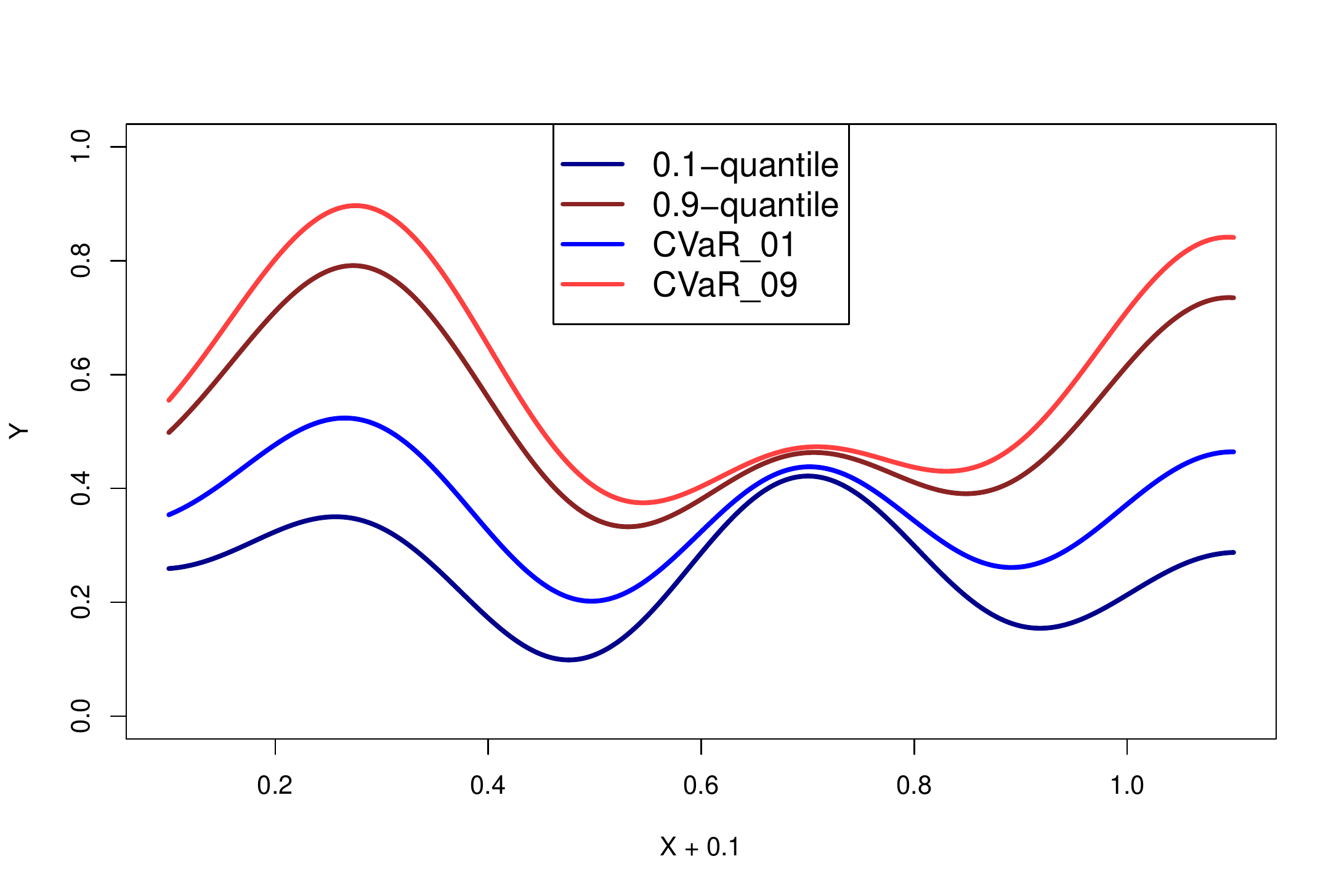}
			\includegraphics[trim={0cm 1cm 0cm 2.0cm},clip,width=7.2cm]{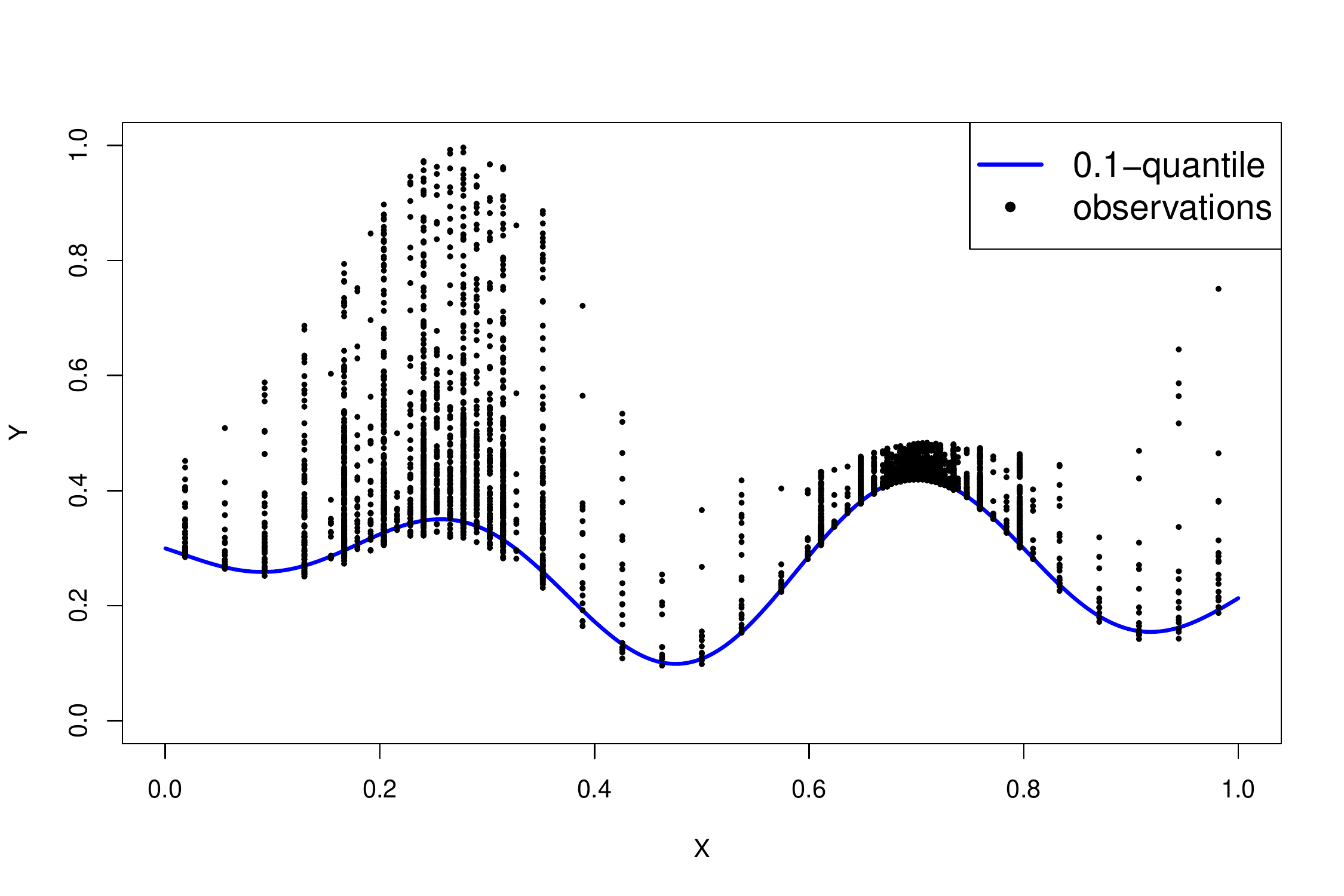}\\
				\includegraphics[trim={0 0cm 0cm 2cm},clip,width=7.2cm]{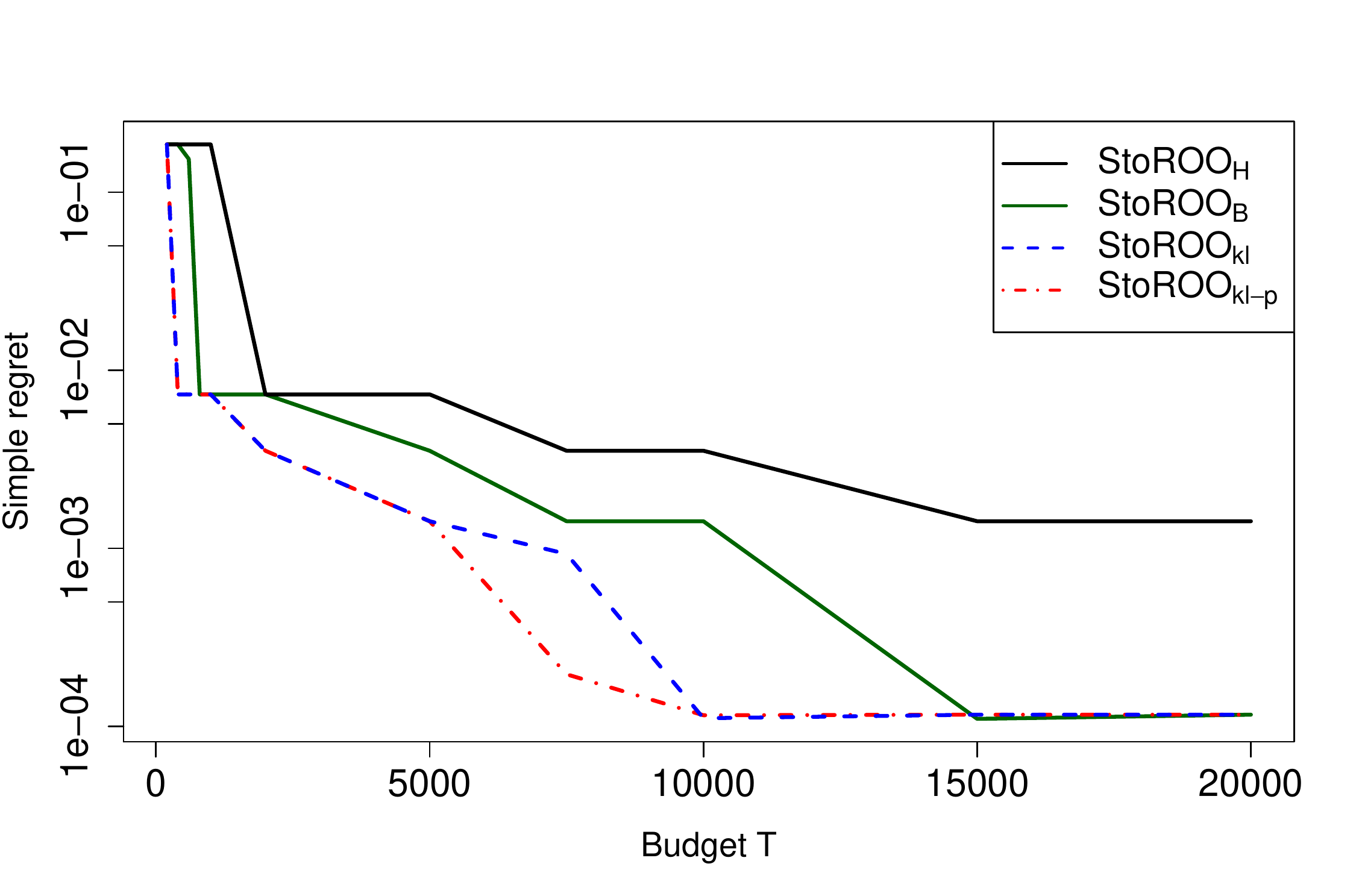}
			\includegraphics[trim={0cm 0cm 0cm 2cm},clip,width=7.2cm]{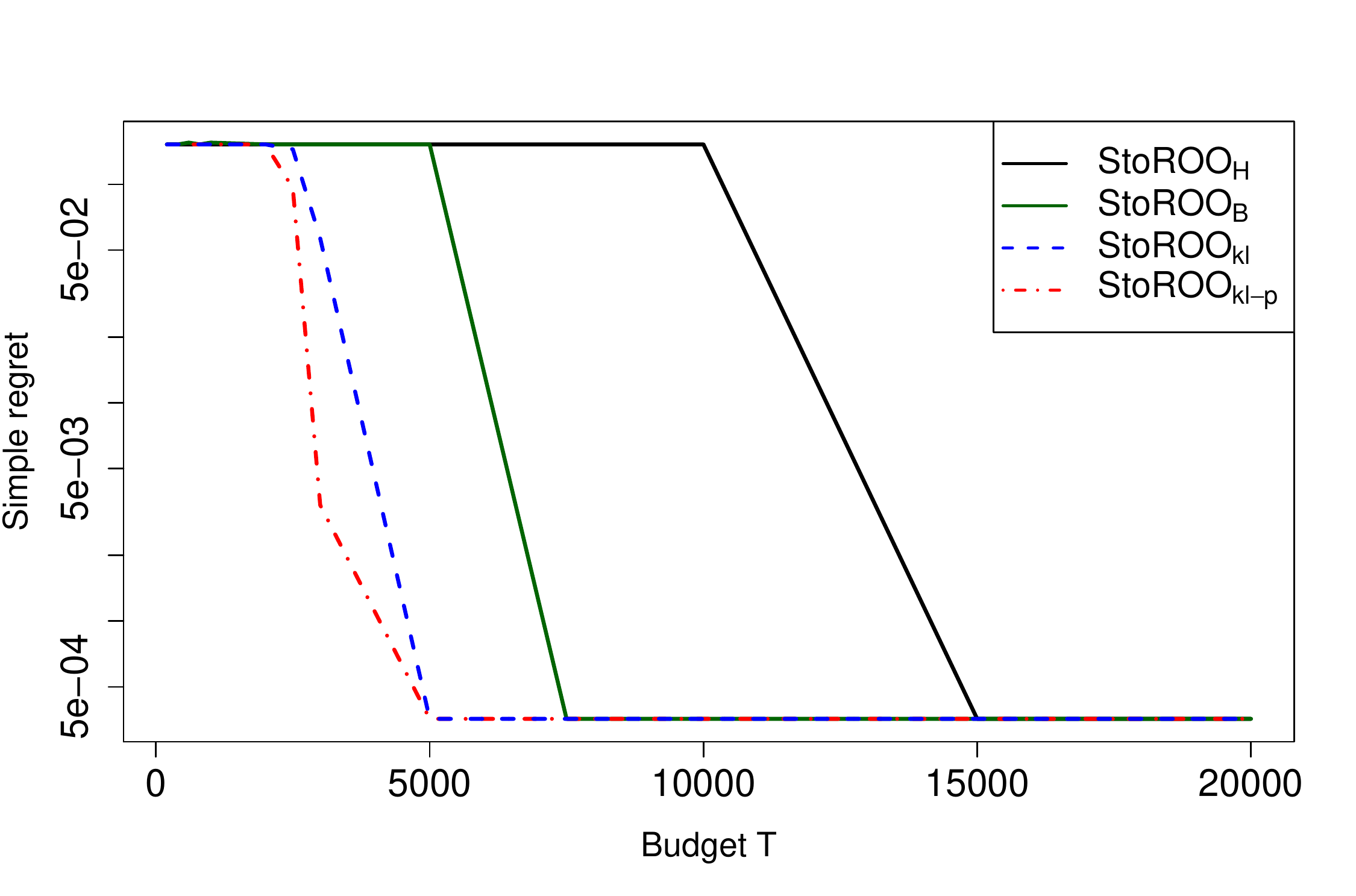}\\
			\includegraphics[trim={0 0cm 0cm 2cm},clip,width=7.2cm]{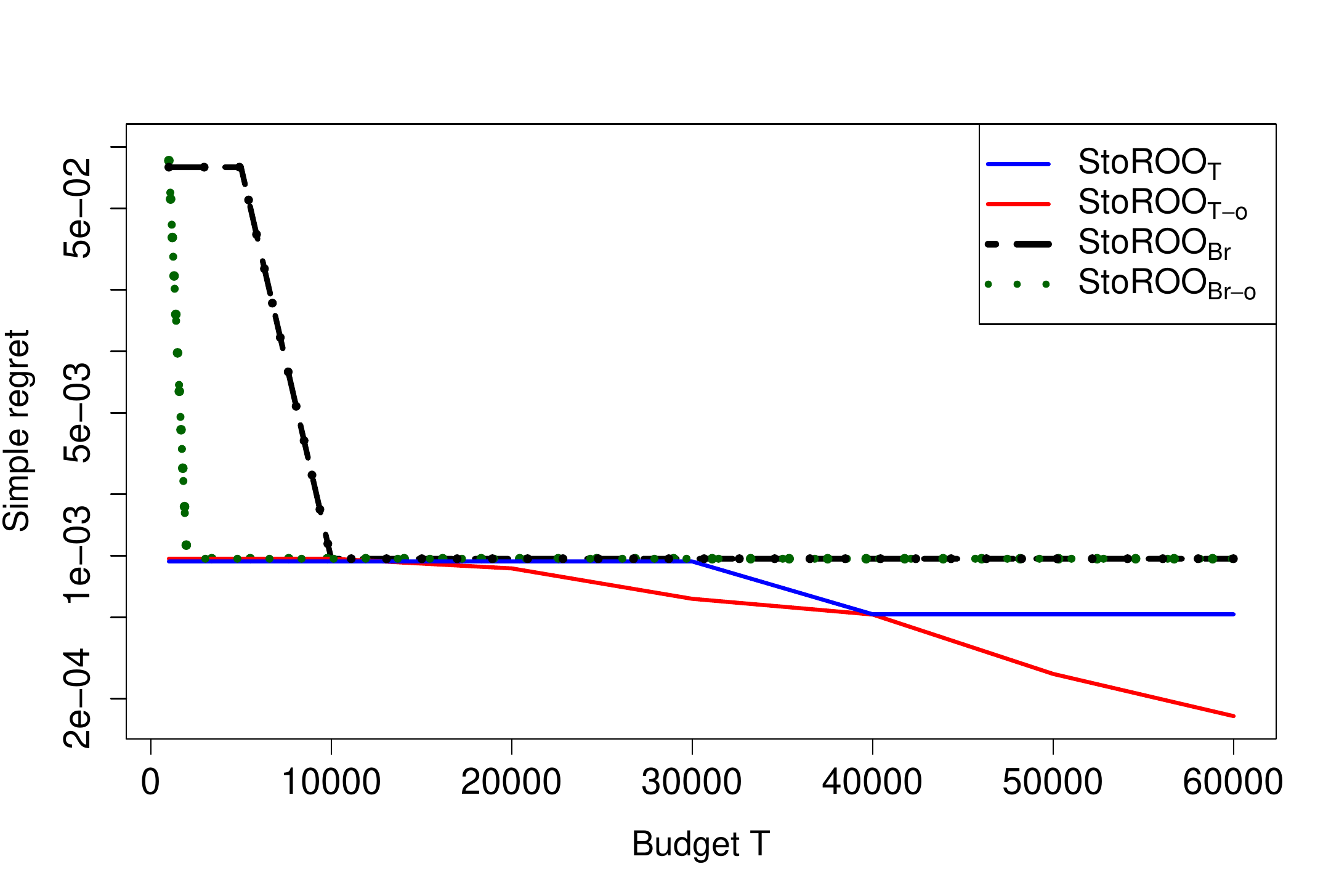}
			\includegraphics[trim={0 0cm 0cm 2cm},clip,width=7.2cm]{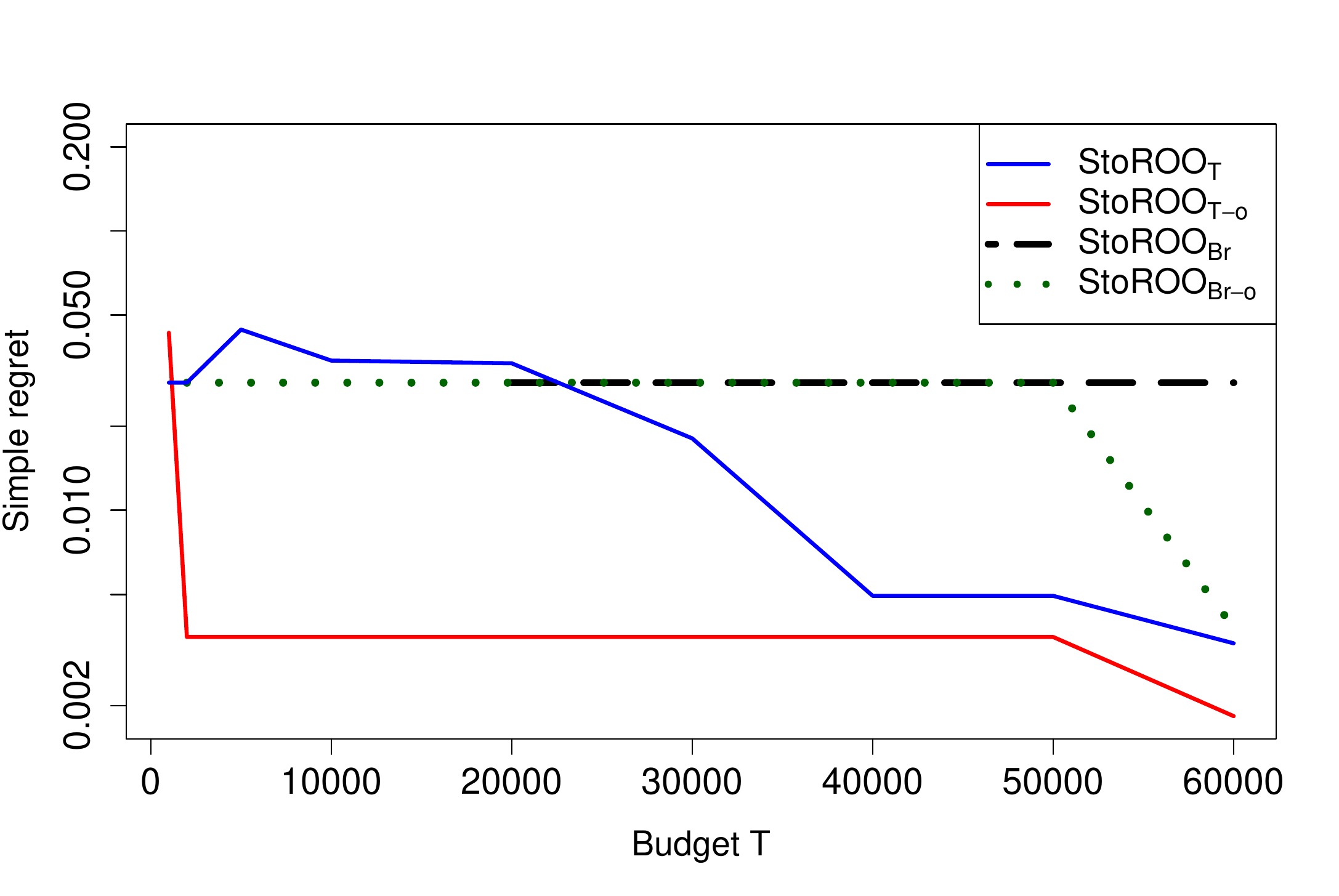}
		\end{tabular}
		\caption{Results for the $\Phi_1$ test function.
		Top left: conditional quantiles and CVaR of $\Phi_1$. Top right: one run of $\StoROO_{\kl}$ for the $0.1$-quantile with $T=5,000$, $\hat{\beta}=12$ and $\hat{\gamma}=1.4$. Middle: evolution of the simple regret for the optimization of the quantile of order $0.1$ (left) and $0.9$ (right). Bottom: evolution of the simple regret for the optimization of the CVaR of order $0.1$ (left) and $0.9$ (right).}\label{quantile_exemples}
	\end{figure}

\begin{figure}
		\begin{tabular}{cc}
			\includegraphics[trim={0 0cm 0cm 2cm},clip,width=7.2cm]{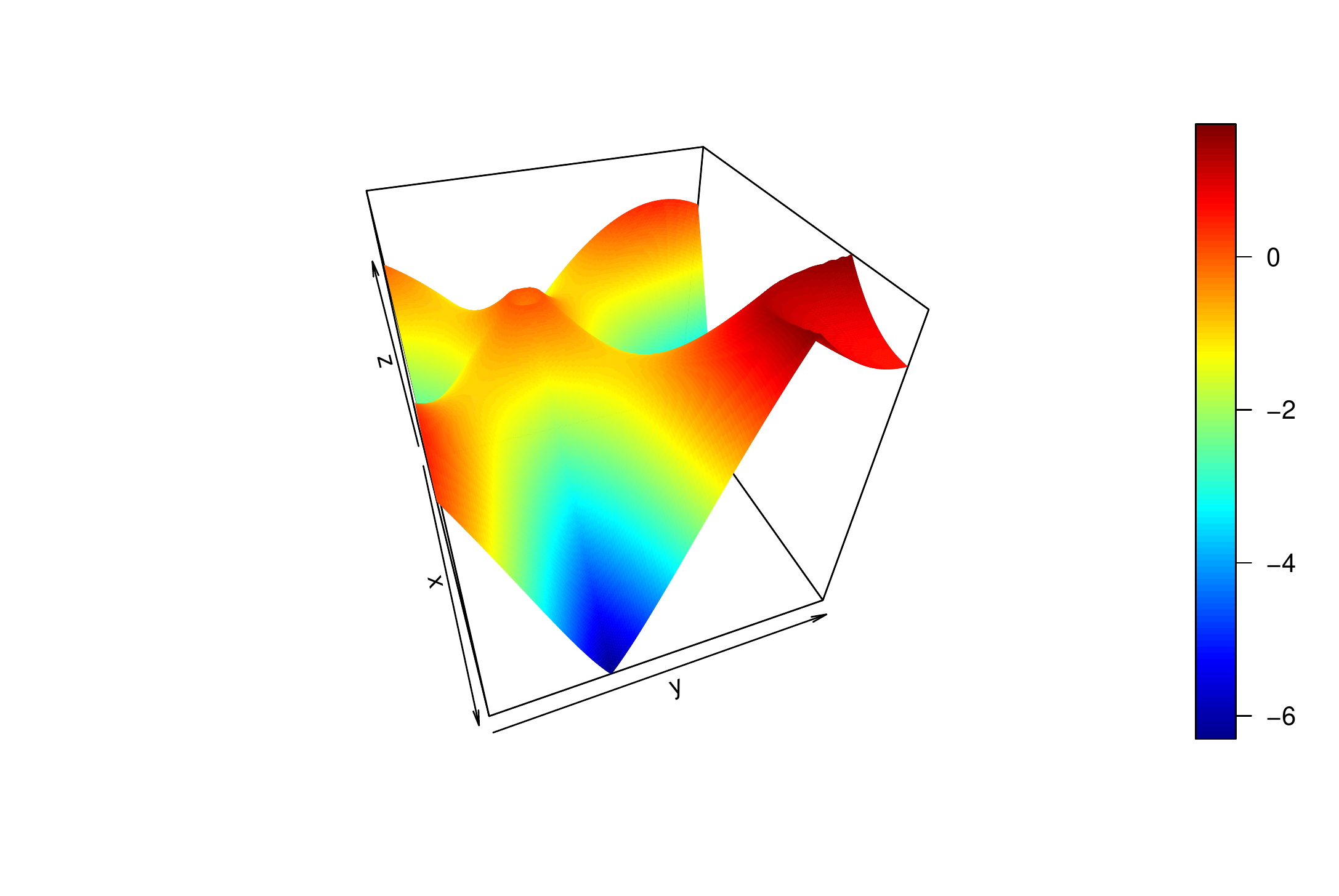}
			\includegraphics[trim={0 0cm 0cm 2cm},clip,width=7.2cm]{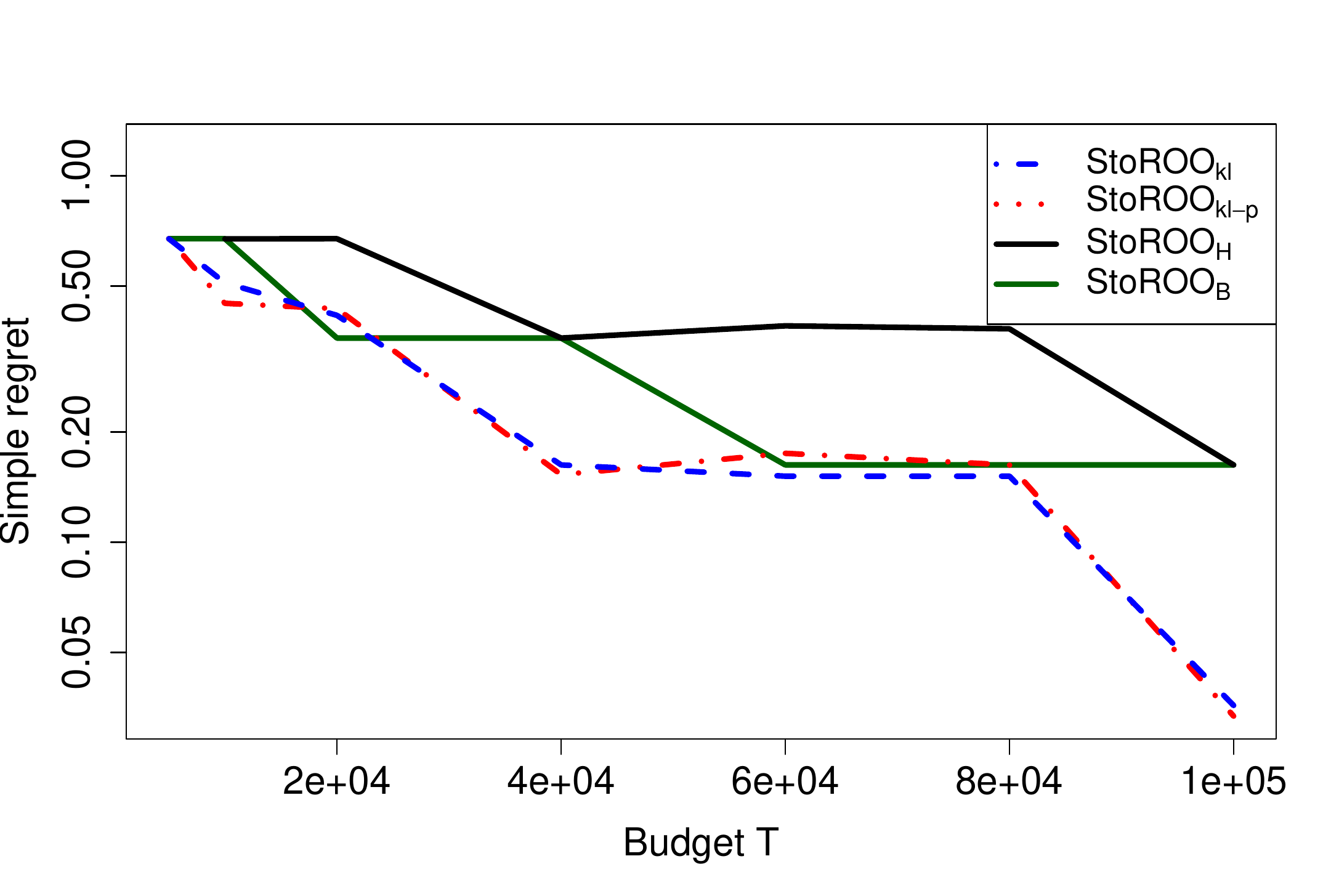}
		\end{tabular}
		\caption{Results for the $\Phi_2$ test function.
		Left: conditional quantile of order $0.1$ of $\Phi_2$, right: simple regret for the optimization of the conditional quantile presented to the left.}\label{2D}
	\end{figure}

	As a test-case, we chose two functions with heteroscedastic noise and local extrema. The first is
	$\Phi_1(x, \cdot)=0.18(\sin(3x)\sin(13x)+1.3)+0.062\zeta(\cdot)\big(\cos(8x-2)+1.2\big),$
	where $\zeta$ is a log-normal random variable of parameters $0$ and $1$ truncated at its $0.95$-quantile (the truncated mass is uniformly reallozcated between $q(0.91)$ and $q(0.95)$). Note that to initialise StoROO not too close from a global optimum, we optimize the quantiles of $\Phi_1$ on $[-0.1,0.9]$ and the CVaR on $[0,1]$.
	Figure \ref{quantile_exemples} (left) shows the shape of the $0.1$ and $0.9$ -quantiles and -CVaR of $\Phi_1$, while Figure \ref{quantile_exemples} (right) shows samples of the $0.1$-quantile. The second test-case is $\displaystyle{\Phi_2(x, \cdot)=\Cr(x)+\zeta(\cdot)|\Cr(x)+1.5\sqrt{x_1^2+x_2^2}|}$, on $[-0.5,1]^2$
	with $$\displaystyle{\Cr(x)=0.1\Big(\Big|\sin(x_1)\sin(x_2)\exp\Big(\big|3-(\sqrt{x_1^2+x_2^2}/\pi)\big|\Big)\Big|+1\Big)^{1.4}}$$ and $\zeta$ a random variable that follows a Cauchy distribution of parameters $(0,0.75)$. Note that for all $x\in\mathcal{X}$, $\Phi_2(x, \cdot)$ is unbounded and it has unbounded moments. Thus we can only apply quantile optimization on $\Phi_2$ based on the strategies developed in the past sections. Figure \ref{2D} (left) shows the shape of the $0.1$-quantile of $\Phi_2$.
	The performance of each version of StoROO is evaluated for different values of $\tau$ and quantified according to the simple regret. 
	In our experiments we fix the values $\hat{\beta}=12$ and $\hat{\gamma}=1.4$ (resp. $\hat{\beta}=2$,  $\hat{\gamma}=0.5$ and $\hat{\beta}=2$, $\hat{\gamma}=0.7$) for the optimization of the quantiles (resp. the CVaR of order $0.1$ and $0.9$) of $\Phi_1$ and $\hat{\beta}=13$ and $\hat{\gamma}=1$ for the optimization of the $0.1$-quantile of $\Phi_2$. % such that the condition (\ref{reg}) is satisfied. 
	Note that these values underestimate the regularity conditions at optimum so that satisfying the condition (\ref{reg}). In addition we fix $K=3^D$ and we choose to expand the nodes into sub-region of equal sizes.
	
	%\begin{figure}
	%	\begin{tabular}{cc}
	%		\includegraphics[width=7.2cm]{regret_cvar_09.pdf}
	%		\includegraphics[width=7.2cm]{cvar_01.pdf}
	%	\end{tabular}
	%	\caption{Evolution of the expectation of the simple regret for the optimization o%f the conditional quantile of $\Phi$: to the left $\tau=0.1$, to the right  $\tau=0.9$.}\label{cvarplot}
	%\end{figure}
	
	%\begin{figure}
	%	\begin{tabular}{cc}
	%		\includegraphics[width=7.2cm]{plot_01_final.pdf}
	%		\includegraphics[width=7.2cm]{plot_09_final.pdf}
	%	\end{tabular}
	%	\caption{Evolution of the expectation of the simple regret for the optimization of the conditional quantile of $\Phi$: to the left $\tau=0.1$, to the right  $\tau=0.9$.}\label{quantile_regret}
	%\end{figure}
	Figure \ref{quantile_exemples} and \ref{2D} report the average of the simple regret over $100$ runs. 
	For both values of $\tau$ all the variants of StoROO have a regret that decreases with the budget. However from our experiments a ranking can be created. 
	For the optimization of the quantile let us firt remark that as bounds are known for $\Phi_1$, for this test case we modified Proposition (\ref{hoeffding}-\ref{union}-\ref{peeling}) by replacing $(-\infty,+\infty$) by $(0,1)$. The less efficient method is $\StoROO_{\h}$. For $\tau=0.9$ its simple regret decreases slower than the three others methods and for $\tau=0.1$ $\StoROO_{\h}$ does not reach the performance of the others variants. To reach a fixed accuracy, $\StoROO_{\h}$ sometimes needs a much larger budget than others variants. For example, on $\Phi_1$, taking $\tau=0.9$, $\StoROO_{\h}$ needs a budget of $15,000$ to reach a simple regret of order $10^{-4}$, while $\StoROO_{\kl}$ and $\StoROO_{\klp}$ need a budget equal to $5,000$.
	Second-to-last is $\StoROO_{\ber}$. Using the maximal budget, on both experiments on $\Phi_1$, this variant reaches the same accuracy as $\StoROO_{\kl}$ and $\StoROO_{\klp}$ but its simple regret decreases slower. For some levels of performance $\StoROO_{\ber}$ needs a much larger budget than $\StoROO_{\kl}$. For example, taking $\tau=0.1$, to reach the value $r_T=10^{-4}$ $\StoROO_{\ber}$ needs a budget of $T=15,000$ while $T=10,000$ is enough for $\StoROO_{\kl}$.
Finally, the most efficient methods are clearly $\StoROO_{\kl}$ and $\StoROO_{\klp}$. 
% Both methods are always at least as good, and often better than $\StoROO_{\h}$ and $\StoROO_{\ber}$. 
The use of a peeling argument (instead of a plain union bound) in $\StoROO_{\klp}$ provides some additional gain over $\StoROO_{\kl}$ on $\Phi_1$ but the effect is negligible on $\Phi_2$.

For the optimization of the CVaR, the variant based on tighter bounds is almost always better than the other and it is independent of the use of oracle bounds. The use of oracle bounds always improves the performance of StoROO and this effect is stronger if the confidence intervals are created with the inequalities of \cite{thomas2019concentration}. Of course, in a real problem the oracle bounds are not known. Nevertheless this result motivates the use of estimators of the minimum and the maximum to estimate the conditional support so that to accelerate convergence.

	\section{Conclusion}
	In this work, we extended StoOO to a generic algorithm applicable to any functional of the reward distribution.
	We proposed a tailored application to the problem of quantile optimization, with four variants: one based on the classical Hoeffding's inequality, one based on Bernstein's inequality, and two others based on Chernoff's inequality. We showed that using Chernoff's inequality to build confidence intervals resulted in a dramatic improvement, both in theory and practice. We also illustrated the ability of StoROO to optimize the CVaR and compared numerically four variants. 
	
	For simplicity, we assumed that the local regularity (or at least, an upper bound) of the target function at the optimum was known to the user.
	However, we believe that it might be  possible to combine our results to the procedure defined in \cite{grill2015black, xuedong2019general} so as to propose an algorithm able to optimize $g$ without
	the knowledge of the smoothness near an optimal point: this is left for future work.
	A second possible extension is to leverage the results proposed here to design an algorithm for the cumulative regret, in the spirit of HOO \cite{bubeck2011x} for example.
% 	\victor{citer HOO ?}\aurelien{oui, et potentiellement d'autres si tu le sens}

\acks{We would like to thank S\'ebastien Gerchinovitz for the discussions and his useful comments.}

% 	\victor{citer HOO ?}\aurelien{oui, et potentiellement d'autres si tu le sens}

\appendix

			\section{Appendix}
			\subsection{Details about the regularity hypothesis}\label{apd:details}
			In the classical setting the Optimized Certainty Equivalent is defined as 
			$$S_u(Y)=\sup_z\Big\{z+\mathds{E}\big(u(Y-z)\big)\Big\},$$
			with $u$ a concave function. Here we assume $u$ is concave and $k$-lipschitzian ($k-\Lip$). Let us consider two random variables $Y_{x_1}$ and $Y_{x_2}$, then
			\begin{align*}
			|S_u(Y_{x_1})-S_u(Y_{x_2})|&=\Big |\sup_z\Big\{z+\mathds{E}\big(u(Y_{x_1}-z)\big)\Big\} - \sup_z\Big\{z+\mathds{E}\big(u(Y_{x_2}-z)\big)\Big\}\Big|\\
			&\leq \sup_z\Big\{\big|\mathds{E}\big(u(Y_{x_1}-z)\big)-\mathds{E}\big(u(Y_{x_2}-z)\big)\big|\Big\}.
			\end{align*}
			Using the \textit{Kantorovich-Rubinstein} representation one obtains
			\begin{align*}
			\sup_z\Big\{\big|\mathds{E}\big(u(Y_{x_1}-z)\big)-\mathds{E}\big(u(Y_{x_2}-z)\big)\big|\Big\} &\leq k\times \mathcal{W}_1(Y_{x_1}-z,Y_{x_2}-z)\\
			&=k \times \mathcal{W}_1(Y_{x_1},Y_{x_2})
			\end{align*}
			with $\mathcal{W}_1$ the Wasserstein distance associated with $p=1$. Thus if $g=S_u$, then a sufficient condition to satisfied (1) is $\mathcal{W}_1(Y_{x*},Y_{x})\leq \frac{\beta}{k} \|x^*-x\|^{\gamma}$, for all $x\in\mathcal{X}$.
			
			To treat the case of the $\cvar$, we use the fact that if $u(z)=\dfrac{\min(z,0)}{1-\tau}$ then we have the equality $S_u=-\cvar$.
			
			In the case of the conditional expectation the same kind of condition can be sufficient. Indeed we have
			$$|\mathds{E}\big(Y_{x_1}\big)-\mathds{E}\big(Y_{x_2}\big)|\leq \sup_{\|f\|\in 1-\Lip}\Big\{\big|\mathds{E}\big(f(Y_{x_1})\big)-\mathds{E}\big(f(Y_{x_2})\big)\big|\Big\} =\mathcal{W}_1(Y_{x_1},Y_{x_2}).$$

			\subsection{Proofs related to the generic analysis of StoROO}\label{apd:first}

			\textbf{Proof of Proposition} \ref{expand}
				
				Let us define $\mathcal{P}_{h^*,j^*}$ the partition containing $x^*$.
				Assume that the partition $\mathcal{P}_{h,j}$ has been selected, thus
				$$\bar{U}_\eta^{h,j}(t)\geq \bar{U}_\eta^{h^*,j^*}(t).$$
				By definition $\bar{U}_\eta^{h^*,j^*}(t)\geq g^*$, thus $\bar{U}_\eta^{h,j}(t)\geq g^*$. 
				Conditionally on $\mathcal{A}_\eta$, $L_\eta^{h,j}(t))\leq g(x_{h,j}(t))$ that implies
				$$g^*-g(x_{h,j})\leq \bar{U}_\eta^{h,j}(t)-L_\eta^{h,j}(t)\leq U_\eta^{h,j}(t)+\bet \delta(h)^{\gamm}-L_\eta^{h,j}(t)\leq 2 \bet \delta(h)^{\gamm}.$$
				Note that the last inequality is obtained because the partition is expanded, which implies that
				$$U(x_{h,j})(t)-L(x_{h,j})(t)\leq  \bet \delta(h)^{\gamm}.$$  
				Finally:
				$$g^*\leq g(x_{h,j})+ 2 \bet \delta(h)^{\gamm},$$
				thus $x_{h,j}$ belongs to $J_h$.

			\textbf{Proof of Proposition} \ref{budg}
				\begin{align*} 
				T=&\sum_{h,j\in \mathcal{T}_T}N_{h,j}(t)
				\leq \sum_{h,j\in \mathcal{T}_T}n_{\eta,h} ~~~~\text{because  $N_{h,j}(t)\leq n_{\eta,h}$}\\
				\leq&\sum_{h'=0}^{\depth(\mathcal{T}_T)-1} K|\mathcal{T}_T \cap J_h|n_{\eta,h'+1}~~~~\text{StoROO has not expanded all the sampled nodes}\\
				\leq&\sum_{h'=0}^{\depth(\mathcal{T}_T)-1}K | J_h|n_{\eta,h'+1}
				=S_{\depth(\mathcal{T}_T)-1}.
				\end{align*} 
				Thus
				$S_{H_\eta}\leq S_{\depth(\mathcal{T}_T)-1} \leq S_{\depth(\mathcal{T}_T)}$
				so
				$H_\eta\leq \depth(\mathcal{T}_T).$
				There is at least an expanded node of depth $H_\eta^*\geq H_\eta$ after a budget $T$ was used.

			\textbf{Proof of Proposition} \ref{regret}
				
				Proposition \ref{expand} implies that the center of an expanded partition is in $J_h$. 
				Proposition \ref{budg} implies that a partition of depth at least $H_\eta^*$ has been expanded. Thus StoROO has expanded a node in $J_{H_\eta^*}$. 
				% 		\victor{il manque un dernier point pour arriver a la borne.}
				At the end of the budget StoROO returns the node having the highest LCB among the nodes that have been expanded and not the deepest node among those that have been expanded. 
					But
					$$g^*-g(x_{h,j})\leq \bar{U}_{H_\eta^*(T),j'}-L_{h,j}\leq \bar{U}_{H_\eta^*(T),j'}-L_{H_\eta^*(T),j'}\leq 2 \bet \delta(H_\eta^*(T))^{\gamm}.$$
					That ensure the node having the highest LCB has the same theoretical regret as the node of maximal depth among those that have been expanded.

				\textbf{Proof of Proposition} \ref{nearopti}
				
					According to the assumption $2$, each cell $\mathcal{P}_{h,j}$ contains a ball of radius $\nu\delta(h)$ centered in $x_{h,j}$ that is a $\ell_{\bet,\gamm}$-ball of radius $\bet(\nu\delta(h))^{\gamm}$  centered in $x_{h,j}$. If $d$ is the $\nu^{\gamm}/2$ near optimality dimension then there is at most $ C[2\bet \delta(h)^{\gamm}]^{-d}$ disjoint $\ell_{\bet,\gamm}$- balls of radius $\bet(\nu\delta(h))^{\gamm}$ inside  $\mathcal{X}_{ 2\bet \delta(h)^{\gamm}}$. Thus if
					$|J_h|=|{x_{h,j}\in \mathcal{X}_{ 2\bet \delta(h)^{\gamm}}}|> C[ \bet\delta(h)^{\gamm}]^{-d}$ this implies there is more than $ C[2 \bet \delta(h)^{\gamm}]^{-d}$ disjoint $\ell_{\bet,\gamm}$ balls of radius $\bet(\nu\delta(h))^{\gamm}$ with center in $\mathcal{X}_{ 2\bet \delta(h)^{\gamm}}$, that is a contradiction.

				\textbf{Proof of Theorem} \ref{main_doo}
				\begin{align*}
				T&\leq \sum_{h=0}^{H^*}K| J_h|n_{\eta,h+1}~~~~~~~~\text{by definition of $H^*$}\\
				&\leq \sum_{h=0}^{H^*}KC[2 \bet \delta(h)^{\gamm}]^{-d}n_{\eta,h+1}~~~~~~~~\text{using Proposition \ref{nearopti}}\\
				&= \sum_{h=0}^{H^*}KC[2 \bet (c\rho^{ h})^{\gamm}]^{-d}n_{\eta,h+1}~~~~~~\text{using the exponential decay of the diameter of the cells}\\
				&\leq\sum_{h=0}^{H^*}KC[2 \bet (c\rho^{ h})^{\gamm}]^{-d}\times \kappa^\alpha\dfrac{\log(T^2/\eta)}{(\bet(c\rho^{ h})^{\gamm})^\alpha}~~~~~~\text{using Definition \ref{conv}}\\
				&= \log(T^2/\eta)\dfrac{KC\kappa^\alpha [2 \bet c^{\gamm}]^{-d}}{\bet c^{\gamm\alpha}} \sum_{h=0}^{H^*} \rho^{h(-d\gamm-\gamm \alpha)}\\
				&= \log(T^2/\eta)\dfrac{KC\kappa^\alpha [2 \bet c^{\gamm}]^{-d}}{\bet c^{\gamm\alpha}}\times\dfrac{\rho^{(H^*+1)(-d\gamm-\gamm \alpha)}-1}{\rho^{-d\gamm-\gamm \alpha}-1}~~~~~~\text{rewriting the sum}\\
				&\leq \dfrac{\log(T^2/\eta)}{(1-\rho^{ d\gamm+\gamm \alpha})}\dfrac{KC\kappa^\alpha [2 \bet c^{\gamm}]^{-d}}{\bet c^{\gamm\alpha}}\times\rho^{H^*(-d\gamm-\gamm \alpha)}\\
				&= \dfrac{\log(T^2/\eta)}{(1-\rho^{ d\gamm+\gamm \alpha})}\dfrac{KC\kappa^\alpha [2 \bet ]^{-d}}{\bet }\times\delta(H^*)^{-d\gamm-\gamm \alpha}.
				\end{align*}
				Finally
				$$\bigg[\dfrac{KC\kappa^\alpha [2 \bet ]^{-d}}{\bet(1-\rho^{ d\gamm+ \gamm \alpha})}\bigg]^{\frac{1}{d\gamm+\gamm \alpha}}\bigg[\dfrac{\log(T^2/\eta)}{T}\bigg]^{\frac{1}{d\gamm+\gamm \alpha}}\geq \delta(H^*).$$
				Using Proposition \ref{regret} we obtain
				$$r_T\leq c_1\Big[\dfrac{ \log(T^2/\eta)}{T}\Big]^{\frac{1}{ \alpha+d}}.$$

			\subsection{Proofs related to the section Optimizing quantiles}

				\textbf{Proof of Proposition} \ref{hoeffding}

		Let us consider the event 
		\begin{eqnarray*}
			\xi_\eta&=&\{\forall~ h\geq 0, \forall~ 0\leq j \leq K^h,\nonumber \forall~ 1\leq t \leq T,\\
			&&\hat{F}_{h,j}^{t}\Big(q_{h,j}(\tau)\Big)\geq\tau+\epsilon_{N_{h,j}(t)}^\eta~\text{or}~ \hat{F}_{h,j}^{t}\Big(q_{h,j}(\tau)\Big)< \tau-\epsilon_{N_{h,j}(t)}^\eta\}.
		\end{eqnarray*}
		\begin{align}
		\mathds{P}\big(\xi_\eta\big)=\mathds{P}\bigg(\forall h\geq 0, \forall 0\leq j \leq K^h,\nonumber \forall 1\leq t \leq T,&~\hat{F}_{h,j}^{t}\Big(q_{h,j}(\tau)\Big)\geq\tau+\epsilon_{N_{h,j}(t)}^\eta ~ \text{or}~,\\ \nonumber&\hat{F}_{h,j}^{t}\Big(q_{h,j}(\tau)\Big)< \tau-\epsilon_{N_{h,j}(t)}^\eta\bigg)\\\nonumber
		\leq~~ \mathds{P}\bigg(\forall h\leq 0, \forall 0\leq j \leq K^h,\nonumber \forall 1\leq t \leq T,&~\hat{F}_{h,j}^{t}\Big(q_{h,j}(\tau)\Big)\geq\tau+\epsilon_{N_{h,j}(t)}^\eta)\bigg)\\
		+~\mathds{P}\bigg(\forall h\geq 0, \forall 0\leq j \leq K^h,\nonumber \forall 1\leq t \leq T,&~\hat{F}_{h,j}^{t}\Big(q_{h,j}(\tau)\Big)< \tau-\epsilon_{N_{h,j}(t)}^\eta\bigg)
		\end{align}
		Define $m\leq T$ the number of nodes expanded throughout the algorithm, define for $1\leq w \leq m$, $\zeta_{w}^s$ as the time when the cell $w$ has been selected for the $s$-th time and define $Y_{w}(\zeta_{w}^s)$ the reward obtained at that time at the point $x_{w}$. Then one can write
		$$\mathds{P}\bigg(\hat{F}_{h,j}^{t}\Big(q_{h,j}(\tau)\Big)\geq\tau+\epsilon_{N_{h,j}(t)}^{\eta,T}\bigg)=\mathds{P}\bigg(\dfrac{1}{N_{h,j}(t)}\sum_{s=1}^{N_{h,j}(t)}\mathds{1}_{Y_{h,j}(\zeta_{h,j}^s)\leq q_{h,j}(\tau)}\geq\tau+\epsilon_{N_{h,j}(t)}^\eta\bigg).$$
		%	\mathds{P}\bigg(\dfrac{1}{N_{h,j}(t)}\sum_{s=1}^{N_{h,j}(t)}Z_{h,j}^\tau(\zeta_{h,j}^s)\geq\tau+\epsilon_{N_{h,j}}^\eta(t)\bigg).
		Using this notation, we have:
		\begin{align}&
		\mathds{P}\bigg(\forall h\geq 0, \forall 0\leq j \leq K^h,\nonumber \forall 1\leq t \leq T,~\hat{F}_{h,j}^{t}\Big(q_{h,j}(\tau)\Big)\geq\tau+\epsilon_{N_{h,j}(t)}^\eta\bigg) \\
		\leq~~ &\mathds{P}\Big(\exists~ 1 \leq w \leq T,~\exists~ 1 \leq u \leq T,~\dfrac{1}{u}\sum_{s=1}^{u}\mathds{1}_{Y_{w}(\zeta_{w}^s)\leq q_{w}(\tau)}\geq\tau+\epsilon_{u}^\eta\Big)\nonumber\\
		\leq~~ &\sum_{w=1}^{T}\sum_{u=1}^{T}\mathds{P}\Big(\dfrac{1}{u}\sum_{s=1}^{u}\mathds{1}_{Y_{w}(\zeta_{w}^s)\leq q_{w}(\tau)}\geq\tau+\epsilon_{u}^\eta\Big)\nonumber
		\end{align}
		By Hoeffding's inequality, if 
		$$\epsilon_{u}^\eta=\sqrt{\dfrac{\log(2T^2/\eta)}{2u}},$$
		we obtain
		$$\mathds{P}\bigg(\forall h\leq 0, \forall 0\leq j \leq K^h,\nonumber \forall 1\leq t \leq T,~\hat{F}_{h,j}^{t}\Big(q_{h,j}(\tau)\Big)\geq\tau+\epsilon_{N_{h,j}(t)}^\eta\bigg)\leq \dfrac{\eta}{2}.$$
		Now using Equation (\ref{ucb1}) we can express this inequality directly in terms of quantiles:
		$$\mathds{P}\bigg(\forall h\leq 0, \forall 0\leq j \leq K^h,\nonumber \forall 1\leq t \leq T,~q_{h,j}(\tau)\geq U_{h,j}^\eta(t)\bigg)\leq \dfrac{\eta}{2}.$$
		Using the same scheme of proof with Inequality (\ref{ucb2}), we obtain: 
		$$\mathds{P}\bigg(\forall h\geq 0, \forall 0\leq j \leq K^h,\nonumber \forall 1\leq t \leq T,~q_{h,j}(\tau)\leq L_{h,j}^\eta(t)\bigg)\leq \dfrac{\eta}{2}, $$
		and hence
		$\mathds{P}\big(\mathcal{A}_\eta\big)=1-\mathds{P}\big(\xi_\eta\big)\geq 1-\eta.$
		
			\textbf{Proof of Proposition} \ref{Massart}

			Without loss of generality let us assume $\tau>0.5$.
			Assume the node $x_{h,j}$ has been sampled $N_{h,j}\geq M_\tau =\max(n_\tau,n_{1-\tau})$ times, with
			$$n_\tau>\dfrac{2\log(2T^2/\eta)}{\tau^2}~~\text{and}~~  n_{1-\tau}>\dfrac{2\log(2T^2/\eta)}{(1-\tau)^2}$$ 
			thus
			$$\tau+2\sqrt{\dfrac{\log(2T^2/\eta)}{2N_{h,j}}}<1 ~\text{and}~ \tau-2\sqrt{\dfrac{\log(2T^2/\eta)}{2N_{h,j}}}>0.$$ 
			That implies 
			$$q_{h,j}\bigg(\tau+2\sqrt{\dfrac{\log(2T^2/\eta)}{2N_{h,j}}}\bigg)<+\infty~\text{and}~ q_{h,j}\bigg(\tau-2\sqrt{\dfrac{\log(2T^2/\eta)}{2N_{h,j}}}\bigg)>-\infty,$$
			and in particular 
			$$U_{h,j}^\eta<+\infty~\text{and}~L_{h,j}^\eta>-\infty.$$
			
			Then define the event 
			$$\mathcal{C}_\eta=\bigcap_{T\geq t\geq 1}\bigcap_{\mathcal{P}_{h,j}\in \mathcal{T}_t}\Big\{q_{h,j}\big(\tau+2\epsilon_{N_{h,j}(t)}^{\eta,T})\big)\geq U_{h,j}^\eta(t) \geq q_{h,j}(\tau)\geq L_{h,j}^\eta(t)\geq q_{h,j}\big(\tau-2\epsilon_{N_{h,j}(t)}^{\eta,T}\big)\Big\},$$
			with $$\epsilon_{N_{h,j}(t)}^{\eta,T}=\sqrt{\dfrac{\log(2T^2/\eta)}{2{N_{h,j}(t)}}}.$$
			Using equivalences (\ref{ucb1}) and (\ref{ucb2}), one can write: 
			\begin{align}
			&q_{h,j}\big(\tau+2\epsilon_{N_{h,j}(t)}^{\eta,T}\big)\geq U_{h,j}^\eta(t) \geq q_{h,j}(\tau)\geq L_{h,j}^\eta(t)\geq q_{h,j}\big(\tau-2\epsilon_{N_{h,j}(t)}^{\eta,T}\big)\nonumber\\ \nonumber&\Leftrightarrow\hat{F}(q_{h,j}(\tau+2\epsilon_{N_{h,j}(t)}^{\eta,T}))\geq\tau+\epsilon_{N_{h,j}(t)}^{\eta,T}>\hat{F}(q_{h,j}(\tau)\geq \tau-\epsilon_{N_{h,j}(t)}^{\eta,T}>\hat{F}(q_{h,j}(\tau+2\epsilon_{N_{h,j}(t)}^{\eta,T})).
			\end{align}
			Thus
			\begin{align*}
			\mathds{P}(\mathcal{C}_\eta)
			&\geq~~1-\mathds{P}(\forall h\geq 0, \forall 0\leq j \leq K^h, \forall 1\leq t \leq T,\sup_{y=q_{\tau},q_{\tau+\epsilon_{N_{h,j}(t)}^{\eta,T}}}|F_{h,j}(y)-\hat{F}_{h,j}^{t}(y)|\geq \epsilon_{N_{h,j}(t)}^{\eta,T}~)\nonumber\\
			&\geq~~1-\mathds{P}(\forall h\geq 0, \forall 0\leq j \leq K^h, \forall 1\leq t \leq T,\sup_{y\in[0,1]}|F_{h,j}(y)-\hat{F}_{h,j}^{t}(y)|\geq \epsilon_{N_{h,j}(t)}^{\eta,T}~).\nonumber\\
			&\text{Using the same notation as in the proof of Proposition $\ref{hoeffding}$, one can write}\\
			&\geq~~1- \sum_{w=1}^{T}\sum_{u=1}^{T}\mathds{P}(\sup_{y\in[0,1]}|F_{w}(y)-\dfrac{1}{u}\sum_{s=1}^{u}\mathds{1}_{Y_{w}(\zeta_{w}^s)\leq q_{w}(\tau)}|\geq \epsilon_{u}^{\eta,T}).\nonumber\\
			\end{align*}
			\text{Now by applying the Massart's inequality to bound} $$\mathds{P}(\sup_{y\in[0,1]}|F_{w}(y)-\sum_{s=1}^{u}\mathds{1}_{Y_{w}(\zeta_{w}^s)\leq q_{w}(\tau)}|\geq \epsilon_{u}^{\eta,T}),$$ \text{ one obtain} $\mathds{P}(\mathcal{C}_\eta)\geq~~1-\eta$. Thus with probability $1-\eta$, we have:
			\begin{equation}
			U_{h,j}^\eta(t)-L_{h,j}^\eta(t)\leq q_{h,j}\bigg(\tau+2\epsilon_{N_{h,j}(t)}^{\eta,T}\bigg)-q_{h,j}\bigg(\tau-2\epsilon_{N_{h,j}(t)}^{\eta,T}\bigg).\label{borne}
			\end{equation}
			Assuming that $q_{h,j}$ is differentiable in $\tau$, by the mean value theorem, we deduce 
			$$q_{h,j}(\tau+2\sqrt{\dfrac{\log(2T^2/\eta)}{2N_{h,j}}})-q_{h,j}(\tau-2\sqrt{\dfrac{\log(2T^2/\eta)}{2N_{h,j}}})\leq 4\sqrt{\dfrac{\log(2T^2/\eta)}{2N_{h,j}}}\max_{\tau'\in [\tau-2\epsilon_{n_{\tau}}^{\eta,T},\tau+2\epsilon_{n_{1-\tau}}^{\eta,T}]} \dfrac{1}{f_{x_{h,j}} \circ F_{x_{h,j}}^{-1}(\tau')}.$$
			Next, using (\ref{borne}) it is possible to write that with probability $1-\eta$:
			$$U_{h,j}^\eta-L_{h,j}^\eta\leq4\sqrt{\dfrac{\log(2T^2/\eta)}{2N_{h,j}}} \dfrac{1}{\bar{f}_{x_{h,j}} }\leq4\sqrt{\dfrac{\log(2T^2/\eta)}{2N_{h,j}}} \dfrac{1}{\min_{x\in\mathcal{X}}\bar{f}(x) }.$$
			We define $n_{\eta,h}'$ as the smallest $n$ such that
			$$4\sqrt{\dfrac{\log(2T^2/\eta)}{2n}} \dfrac{1}{\inf_{x\in\mathcal{X}}\bar{f}(x) }\leq \bet\delta(h)^{\gamm},$$
			that is
			$$n_{\eta,h}'=\log(2T^2/\eta)\bigg(\dfrac{2\sqrt{2}}{\bet\delta(h)^{\gamm}\min_{x\in\mathcal{X}}\bar{f}(x) }\bigg)^2.$$
			A proper $n_{\eta,h}$ has to verify 
			$$n_{\eta,h}\geq M_\tau ~\text{and}~n_{\eta,h}\geq\log(2T^2/\eta)\bigg(\dfrac{2\sqrt{2}}{\bet\delta(h)^{\gamm}\min_{x\in\mathcal{X}}\bar{f}(x) }\bigg)^2.$$
			To satisfy this constraint we define 
			\begin{align*}
			n_{\eta,h}=&~\log(2T^2/\eta)\Bigg(\dfrac{\sqrt{8\min(1-\tau,\tau)^2+ 4\big(\bet\diam(\mathcal{X})^{\gamm}\min_{x\in\mathcal{X}}\bar{f}(x)\big)^2}}{\bet\delta(h)^{\gamm} \min_{x\in\mathcal{X}}\bar{f}(x)\min(1-\tau,\tau) }\Bigg)^2\\
			\geq&~\log(2T^2/\eta)\Bigg(\bigg(\dfrac{2\sqrt{2}}{\bet\delta(h)^{\gamm}\min_{x\in\mathcal{X}}\bar{f}(x) }\bigg)^2+\bigg(\dfrac{2}{\min(1-\tau,\tau)}\bigg)^2\Bigg)\\
			=&~ n_{\eta,h}'+M_\tau.
			\end{align*}
			To conclude the whole proof, since $\mathcal{C}_\eta\subset\mathcal{A}_\eta\cap\mathcal{B}_\eta$, we obtain $\mathds{P}(\mathcal{A}_\eta\cap\mathcal{B}_\eta)\geq1-\eta$.

		\textbf{Proof of Proposition} \ref{unionb}
				
				Let $Y_1,\cdots,Y_n$ be $n$ $i.i.d.$ random variables bounded by the interval $[0,1]$. Define $\hat{F}^n(q(\tau))=\frac{1}{n}\sum_{i=1}^n \mathds{1}_{Y_i\leq q(\tau)}$. For $x>\tau$ the Bernstein's inequality gives
				$$\mathds{P}(|\hat{F}^n(q(\tau))-\tau|>\epsilon)\leq 2\exp\bigg(\dfrac{n\epsilon^2}{2\tau(1-\tau)+2\epsilon/3}\bigg).$$
				Let us consider the event 
				\begin{eqnarray*}
					\xi_\eta&=&\{\forall~ h\geq 0, \forall~ 0\leq j \leq K^h,\nonumber \forall~ 1\leq t \leq T,\\
					&&\hat{F}_{h,j}^{t}\Big(q_{h,j}(\tau)\Big)\geq\tau+\epsilon_{N_{h,j}(t)}^{\eta,T}~\text{or}~ \hat{F}_{h,j}^{t}\Big(q_{h,j}(\tau)\Big)< \tau-\epsilon_{N_{h,j}(t)}^{\eta,T}\}.
				\end{eqnarray*}
				Using the same lines as in the proof of Proposition \ref{hoeffding} we have
				\begin{align}
				\mathds{P}(\xi_\eta) 
				\leq~~ &\sum_{w=1}^{T}\sum_{u=1}^{T}\mathds{P}\Big(|\dfrac{1}{u}\sum_{s=1}^{u}\mathds{1}_{Y_{w}(\zeta_{w}^s)\leq q_{w}(\tau)}-\tau|>\epsilon_{u}^{\eta,T}\Big)\nonumber\\
				&\text{then applying the Bernstein's inequality we obtain}\nonumber\\
				\leq~~&\sum_{w=1}^{T}\sum_{u=1}^{T}2\exp\bigg(-\dfrac{u{\epsilon_{N_{h,j}(t)}^{\eta,T}}^2}{2\tau(1-\tau)+2\epsilon_{N_{h,j}(t)}^{\eta,T}/3}\bigg)\label{bern}.
				\end{align}
				By now the goal is to find $\epsilon_{N_{h,j}(t)}^{\eta,T}>0$ such that 
				$$\dfrac{u{\epsilon_{N_{h,j}(t)}^{\eta,T}}^2}{2\tau(1-\tau)+2\epsilon_{N_{h,j}(t)}^{\eta,T}/3}=\log(2T^2/\eta).$$
				Finding such $\epsilon_{N_{h,j}(t)}^{\eta,T}$ can be easily done because it is a square of a second order polynomial. The result is
				$$\epsilon_{N_{h,j}(t)}^{\eta,T}=\dfrac{\log(2T^2/\eta)}{3u}\bigg(1+\sqrt{1+\dfrac{18u\tau(1-\tau)}{\log(2T^2/\eta)}}\bigg).$$
				Plugging the value of $\epsilon_{N_{h,j}(t)}^{\eta,T}$ inside (\ref{bern}) concludes the proof.

				\textbf{Proof of Proposition} \ref{union}
		
		\textbf{Step 1: bounds on $\hat{F}^n(q(\tau))$ for a i.i.d sample}\\\\
		Let $Y_1,\cdots,Y_n$ be $n$ $i.i.d.$ random variables bounded by the interval $[0,1]$. Define $\hat{F}^n(q)=\frac{1}{n}\sum_{i=1}^n \mathds{1}_{Y_i\leq q}$. For $x>\tau$ the Chernoff's inequality gives
		$$\mathds{P}(\hat{F}^n(q(\tau))\geq x)\leq \exp(-n \kl(x,\tau)).$$
		Let $\tau^+>\tau$ be the value such that $\kl(\tau^+,\tau)= \frac{\log(2/\eta)}{n}$, then for all $x\geq \tau^+$:
		$$\mathds{P}(\hat{F}^n(q(\tau))\geq x)\leq \mathds{P}(\hat{F}^n(q(\tau))\geq \tau^+)\leq\exp(n\frac{\log(2/\eta)}{n})=\dfrac{\eta}{2}.$$
		Now let us define the candidate for the UCB of a i.i.d sample:
		$$U(n)=\min\big\{q,~~\hat{F}^n(q)\geq\tau~\text{and}~n \kl(\hat{F}^n(q),\tau)\geq \log(2/\eta)\big\},$$ and let us remark that 
		\begin{equation}
		\hat{F}^n(U(n))\leq\hat{F}^n(q(\tau))\Leftrightarrow \tau\leq \hat{F}^n(q(\tau)) ~~\text{and}~~ \kl(\hat{F}^n(q(\tau)),\tau)\geq \frac{\log(2/\eta)}{n},\label{equi}
		\end{equation}
		thus 
		\begin{align*}
		\mathds{P}(\hat{F}^n(U(n))\leq\hat{F}^n(q(\tau)))=&	\mathds{P}(\tau\leq \hat{F}^n(q(\tau)) ~~\text{and}~~ \kl(\hat{F}^n(q(\tau)),\tau)\geq \frac{\log(2/\eta)}{n})\\
		\leq&\mathds{P}(\hat{F}^n(q(\tau))\geq \tau^+)\leq \dfrac{\eta}{2}.
		\end{align*}

		For $x<\tau$ let us introduce
		$$L(n)=\max\big\{q,~~\hat{F}^n(q)\leq\tau~\text{and}~n \kl(\hat{F}^n(q),\tau)\geq \log(2/\eta)\big\},$$one proves in the same way 
		\begin{align*}
		\mathds{P}(\hat{F}^n(L(n))>\hat{F}^n(q(\tau)))\leq \dfrac{\eta}{2}.
		\end{align*}
		
		\textbf{Step 2: Double union bound}\\\\
		Let us consider the event 
		\begin{eqnarray*}
			\xi_\eta&=&\Big\{\forall~ h\geq 0, \forall~ 0\leq j \leq K^h,\nonumber \forall~ 1\leq t \leq T,\\
			&&\hat{F}_{h,j}^{t}\big(q_{h,j}(\tau)\big)\geq\hat{F}_{h,j}^{t}(U_{h,j}^\eta)~\text{or}~ \hat{F}_{h,j}^{t}\big(q_{h,j}(\tau)\big)< \hat{F}_{h,j}^{t}(L_{h,j}^\eta)\Big\}.
		\end{eqnarray*}
		\begin{align}
		\mathds{P}\big(\xi_\eta\big)\leq~~ \mathds{P}\bigg(\forall h\leq 0, \forall 0\leq j \leq K^h,\nonumber \forall 1\leq t \leq T,&~\hat{F}_{h,j}^{t}\big(q_{h,j}(\tau)\big)\geq\hat{F}_{h,j}^{t}(U_{h,j}^\eta)\bigg)\\
		+~\mathds{P}\bigg(\forall h\geq 0, \forall 0\leq j \leq K^h,\nonumber \forall 1\leq t \leq T,&~\hat{F}_{h,j}^{t}\big(q_{h,j}(\tau)\big)< \hat{F}_{h,j}^{t}(L_{h,j}^\eta)\bigg)
		\end{align}
		Following the notation of the proof of Proposition \ref{hoeffding} we have
		\begin{align}&
		\mathds{P}\bigg(\forall h\geq 0, \forall 0\leq j \leq K^h,\nonumber \forall 1\leq t \leq T,~\hat{F}_{h,j}^{t}\big(q_{h,j}(\tau)\big)\geq\hat{F}_{h,j}^{t}(U_{h,j}^\eta)\bigg) \\
		\leq~~ &\mathds{P}\Big(\exists~ 1 \leq w \leq T,~\exists~ 1 \leq u \leq T,~\sum_{s=1}^{u}\mathds{1}_{Y_{w}(\zeta_{w}^s)\leq q_{w}(\tau)}\geq\sum_{s=1}^{u}\mathds{1}_{Y_{w}(\zeta_{w}^s)\leq U_{w}^\eta}\Big)\nonumber\\
		\leq~~ &\sum_{w=1}^{T}\sum_{u=1}^{T}\mathds{P}\Big(\sum_{s=1}^{u}\mathds{1}_{Y_{w}(\zeta_{w}^s)\leq q_{w}(\tau)}\geq\sum_{s=1}^{u}\mathds{1}_{Y_{w}(\zeta_{w}^s)\leq U_{w}^\eta}\Big).\nonumber
		\end{align}
		\text{Using the equivalence (\ref{equi}), the probability can be reformulated as }
		\begin{align}
		=~~ &\sum_{w=1}^{T}\sum_{u=1}^{T}\mathds{P}\Big(\tau\leq \hat{F}^u(q(\tau)) ~~\text{and}~~ \kl(\hat{F}^u(q(\tau)),\tau)\geq \frac{\log(2T^2/\eta)}{u}\Big).\nonumber
		\end{align}
		Now using Chernoff's inequality we obtain
		\begin{align}
		&\mathds{P}\bigg(\forall h\geq 0, \forall 0\leq j \leq K^h,\nonumber \forall 1\leq t \leq T,~\hat{F}_{h,j}^{t}\big(q_{h,j}(\tau)\big)\geq\hat{F}_{h,j}^{t}(U_{h,j}^\eta)\bigg)\\
		&\leq~~\sum_{w=1}^{T}\sum_{u=1}^{T}\exp(-u\dfrac{\log(2T^2/\eta)}{u})=\eta/2.\nonumber
		\end{align}
		By equivalence (\ref{ucb1}) this implies that, $\forall h\geq 0, \forall 0\leq j \leq K^h,\nonumber \forall 1\leq t \leq T$, with probability at least $\eta/2$,
		$U_{h,j}^\eta(t)~\leq~q_{h,j}(\tau).$
		Using the same lines one can show 
		\begin{align}
		\mathds{P}\bigg(\forall h\geq 0, \forall 0\leq j \leq K^h,\nonumber \forall 1\leq t \leq T,&~\hat{F}_{h,j}^{t}\Big(q_{h,j}(\tau)\Big)< \hat{F}_{h,j}^{t}(L)\bigg)\leq \eta/2,
		\end{align}
		By equivalence (\ref{ucb2}) this implies that, $\forall h\geq 0, \forall 0\leq j \leq K^h,\nonumber \forall 1\leq t \leq T$, $L_{h,j}^\eta(t)> q_{h,j}(\tau)$ with probability at least $\eta/2$. Putting this two probabilities together prove the result.

			\textbf{Proof of Proposition} \ref{peeling}
				
		Define 
		$$\tilde{S}_{h,j}^{\tau}(n)=\sum_{i=1}^n \mathds{1}_{Y_{h,j}(i)\leq q_{h,j}(\tau)}.$$
		\textbf{Step 1: Martingale}
		For every $\lambda\in \mathds{R}$, let $\phi_\tau(\lambda)=\log\mathds{E}[\exp(\lambda \mathds{1}_{Y_{h,j}(1)\leq q_{h,j}(\tau))}]$. Let $W_0^{\lambda}=1$ and for $n\geq 1$,
		$$W_n^\lambda=\exp(\lambda \tilde{S}_{h,j}^{\tau}(n)-n\phi_\tau(\lambda)).$$
		$(W_n^\lambda)_{n\geq0}$ is a martingale relative to $(\mathcal{F}_n)_{n\geq 0}$. In fact,
		
		\begin{align*}
		\mathds{E}\Big[\exp\Big(\lambda\{\tilde{S}_{h,j}^{\tau}(n+1)-\tilde{S}_{h,j}^{\tau}(n)\}\Big)|\mathcal{F}_n\Big]=&\mathds{E}\Big[\exp(\lambda X_{n+1})|\mathcal{F}_n\Big]\\
		=&\exp\Big(\log\mathds{E}[\exp(\lambda X_1]\Big)\\
		=&\exp\Big(\{(n+1)-n\}\phi_\mu(\lambda)\Big)
		\end{align*}
		
		That is equivalent to
		$$\mathds{E}\Big[\exp\Big(\lambda\{\tilde{S}_{h,j}^{\tau}(n+1)-\tilde{S}_{h,j}^{\tau}(n)\}\Big)|\mathcal{F}_n\Big]= \exp\Big(\lambda S_n-n \phi_\mu(\lambda)\Big).$$
		\textbf{Step 2: Peeling}
		Let us devide the interval $\{1,\cdots,T\}$ into \textit{slices} $\{t_{k-1}+1,\cdots,t_{k}\}$ of geometric increasing size. We may assume that $\delta>1$, since otherwise the bound is trivial. Take $\xi=1/(1-\delta_\eta(T))$, let $t_0=0$ and for all $k\in\mathds{N}^*$, let $t_k=\lfloor(1+\xi)^k\rfloor$.
		\begin{align}&
		\mathds{P}\bigg(\forall h\geq 0, \forall 0\leq j \leq K^h,\nonumber \forall~ 1\leq t \leq T,~U_{h,j}^\eta(t)\leq q_{h,j}(\tau)\bigg) \\
		\leq~~ &\mathds{P}\Big(\exists~h\geq 0, \exists~ 0\leq j \leq K^h,~\exists~ 1 \leq t \leq T,~U_{h,j}^\eta (t)\leq q_{h,j}(\tau)\Big).\nonumber
		\end{align}
		Define $m\leq T$ the number of nodes expanded throughout the algorithm, thus for $1~\leq~w~\leq~m$, it is possible to rewrite the last probability as
		\begin{align}&
		\mathds{P}\Big(\exists~ 1 \leq w \leq T,~\exists~ 1 \leq n \leq T,~U_{w}^\eta (n)\leq q_{w}(\tau)\Big)\nonumber\\
		\leq~~ &\sum_{w=1}^{T}\mathds{P}\Big(~\exists~ 1 \leq k \leq D,~\exists~t_{k-1}< n\leq t_{k}~~ \text{and}~~ U_{w}^\eta (n)\leq q_{w}(\tau)\Big)\nonumber~~~~\text{with}~~D=\frac{\log(T)}{\log(1+\eta)}\\
		\leq~~ &\sum_{w=1}^{T}\sum_{k=1}^{D}\mathds{P}\Big(A_k\Big),\nonumber
		\end{align}
		with $$A_k=\big\{~\exists~t_{k-1}< n\leq t_{k}~~ \text{and}~~ U_{w}^\eta (n)\leq q_{w}(\tau)\big\}.$$
		Observe that $U_{w}^\eta (n)\leq q_{w}(\tau)$ if and only if $\dfrac{1}{n}\sum_{s=1}^{u}\mathds{1}_{Y_{w}(\zeta_{w}^s)\leq U_{w}^\eta}\leq \dfrac{1}{n}\tilde{S}_{w}^{\tau}(n)$ and
		$$\dfrac{1}{n}\sum_{s=1}^{u}\mathds{1}_{Y_{w}(\zeta_{w}^s)\leq U_{w}^\eta}\leq \frac{\tilde{S}_{w}^{\tau}(n)}{n}\Leftrightarrow \tau\leq \frac{\tilde{S}_{w}^{\tau}(n)}{n} ~~\text{and}~~ \kl(\frac{\tilde{S}_{w}^{\tau}(n)}{n},\tau)\geq \delta_\eta(T)+\frac{1}{n}.$$
		Define $\delta=\delta_\eta(T)+1/n $, let $s$ be the smallest integer such that $\delta/(s+1)\leq \kl(1,\tau)$; if $n\leq s$, then $n~\kl(\frac{\tilde{S}_{w}^{\tau}(n)}{n},\tau)~\leq ~s \kl(\frac{\tilde{S}_{w}^{\tau}(n)}{n},\tau) \leq s \kl(1,\tau)<\delta$ thus $\mathds{P}(U(n)<q(\tau))=0$. Thus for all $k$ such that $t_k\geq s$, we obtain $\mathds{P}(A_k=0)$.
		For $k$ such that $t_k>s$, let $\tilde{t}_{k-1}=\max\{t_{k-1},s\}.$ Let $x\in ]\tau,1[$ be such that $\kl(x,\tau)=\delta/n$ and let $\lambda(x)=\log(x(1-\tau))-\log(\tau(1-x))>0$, so that $\kl(x,\tau)=\lambda(x) x-(1-\tau+\tau\exp(\lambda(x))).$ Consider $z$ such that $z>\tau$ and $\kl(z,\tau)=\delta/(1+\xi)^k$.
		
		Observe that
		\begin{itemize}
			\item if $n > \tilde{t}_{k-1}$, then 
			$$\kl(z,\tau)=\dfrac{\delta}{(1+\xi)^k}\geq \dfrac{\delta}{(1+\xi)n};$$
			\item if $n\leq t_k$, then as 
			$$\kl\big(\frac{\tilde{S}_{w}^{\tau}(n)}{n},\tau\big)>\dfrac{\delta}{n}>\dfrac{\delta}{(1+\xi)^k}=\kl(z,\tau),$$
		\end{itemize}
		it holds that:
		$$\tau\leq\frac{\tilde{S}_{w}^{\tau}(n)}{n} ~~\text{and}~~ \kl(\frac{\tilde{S}_{w}^{\tau}(n)}{n},\tau)\geq \frac{\delta}{n} \Rightarrow \frac{\tilde{S}_{w}^{\tau}(n)}{n}\geq z.$$
		Hence on the event $\{\tilde{t}_{k-1}<n<t_k\}\cap\{\tau\leq \frac{\tilde{S}_{w}^{\tau}(n)}{n}\}\cap\{\kl(\frac{\tilde{S}_{w}^{\tau}(n)}{n},\tau)\geq \frac{\delta}{n} \}$ it holds that
		$$\lambda(z)\frac{\tilde{S}_{w}^{\tau}(n)}{n}\geq \lambda(z)z-\phi_{\tau}(\lambda(z))=\kl(z,\tau)\geq \dfrac{\delta}{(1+\xi)n}.$$
		\textbf{Step 3: Putting everything together}
		$$\{\tilde{t}_{k-1}<n<t_k\}\cap\{\tau\leq \frac{\tilde{S}_{w}^{\tau}(n)}{n}\}\cap\{\kl(\frac{\tilde{S}_{w}^{\tau}(n)}{n},\tau)\geq \frac{\delta}{n} \}$$
		\begin{align*}
		\subset& \{\lambda(z)\frac{\tilde{S}_{w}^{\tau}(n)}{n}-\phi_{\tau}(\lambda(z))\geq \dfrac{\delta}{n(1+\xi)}\}\\
		\subset&\{\lambda(z)S_w(n)-n\phi_{\tau}(\lambda(z))\geq \dfrac{\delta_\eta(T)}{(1+\xi)}\}\\
		\subset&\{W_n^{\lambda(z)} >\exp( \dfrac{\delta_\eta(T)}{(1+\xi)})\}.
		\end{align*}
		As $(W_n^\lambda)_{n\geq0}$ is a martingale, $\mathds{E}[W_n^{\lambda(z)}]\leq \mathds{E}[W_0^{\lambda(z)}]=1$. Thus the Doob's inequality for martingales provides:
		%$$\mathds{P}(\{\tilde{t}_{k-1}<n<t_k\}\cap\{\tau\leq \hat{F}_w^n(q(\tau))\}\cap\{\kl(\hat{F}_w^n(q(\tau)),\tau)\geq \frac{\delta}{n} \})$$
		$$\mathds{P}\Bigg(\sup_{\tilde{t}_{k-1}<n<t_k}W_n^{\lambda(z)}>\exp\Big(\dfrac{\delta_\eta(T)}{1+\xi}\Big)\Bigg)\leq \exp\Big(-\dfrac{\delta_\eta(T)}{1+\xi}\Big)$$
		Finally
		$$\sum_{w=1}^{T}\sum_{k=1}^{D}\mathds{P}\Big(~\exists~t_{k-1}< n\leq t_{k}~~ \text{and}~~ U_{w}^\eta (n)\leq q_{w}(\tau)\Big)\leq T D  \exp(-\dfrac{\delta_\eta(T)}{(1+\xi)}).$$
		But as $\xi=1/(\delta_\eta(T)-1)$, $D=\Big\lceil\dfrac{\log(T)}{\log(1+1/(\delta_\eta(T)+1))}\Big\rceil$ and as long as $$\log(1+1/(\delta_\eta(T)-1))\geq 1/\delta_\eta(T),$$ we obtain:
		$$\mathds{P}(\mathcal{A}^c)\leq T \Big\lceil\dfrac{\log(T)}{\log(1+1/(\delta_\eta(T)+1))}\Big\rceil\exp(-\delta_\eta(T) +1)\leq T e \lceil \delta_\eta(T) \log(T)\rceil \exp(-\delta_\eta(T))\leq \eta/2.$$
		Using the same lines for the LCB concludes the proof.

			\subsection{Proofs related to the section Optimizing CVaR}
			
			\textbf{Proof of Proposition} \ref{Brown}

		Let us consider the event 
		\begin{eqnarray*}
			\xi_\eta&=&\Big\{\forall~ h\geq 0, \forall~ 0\leq j \leq K^h,\nonumber \forall~ 1\leq t \leq T,\\
			&&\widehat{\cvar}^{t}(Y_{x_{h,j}})\geq\cvar(Y_{x_{h,j}})+\tilde{\epsilon}_{N_{h,j}(t)}^\eta~\text{or}~ \widehat{\cvar}^{t}(Y_{x_{h,j}})\leq \cvar(Y_{x_{h,j}})-\epsilon_{N_{h,j}(t)}^\eta\Big\}.
		\end{eqnarray*}
		\begin{align}
		\mathds{P}\big(\xi_\eta\big)=\mathds{P}\bigg(\forall h\geq 0, \forall 0\leq j \leq K^h,\nonumber \forall 1\leq t \leq T,&~\widehat{\cvar}^{t}(Y_{x_{h,j}})\geq\cvar(Y_{x_{h,j}})+\tilde{\epsilon}_{N_{h,j}(t)}^\eta~ \text{or}~,\\ \nonumber&\widehat{\cvar}^{t}(Y_{x_{h,j}})\leq \cvar(Y_{x_{h,j}})-\epsilon_{N_{h,j}(t)}^\eta\bigg)\\
		\leq~~ \mathds{P}\bigg(\forall h\leq 0, \forall 0\leq j \leq K^h, \forall 1\leq t \leq\label{ff} T,&~\widehat{\cvar}^{t}(Y_{x_{h,j}})\geq\cvar(Y_{x_{h,j}})+\tilde{\epsilon}_{N_{h,j}(t)}^\eta\bigg)\\
		+~\mathds{P}\bigg(\forall h\geq 0, \forall 0\leq j \leq K^h, \forall 1\leq t \leq T,&~\widehat{\cvar}^{t}(Y_{x_{h,j}})\leq \cvar(Y_{x_{h,j}})-\epsilon_{N_{h,j}(t)}^\eta\bigg)\label{ss}
		\end{align}
		First let us consider (\ref{ff}):
		\begin{align}&
		\mathds{P}\bigg(\forall h\geq 0, \forall 0\leq j \leq K^h,\nonumber \forall 1\leq t \leq T,~\widehat{\cvar}^{t}(Y_{x_{h,j}})\geq\cvar(Y_{x_{h,j}})+\tilde{\epsilon}_{N_{h,j}(t)}^\eta\bigg) \\
		\leq~~ &\mathds{P}\Big(\exists~ 1 \leq w \leq T,~\exists~ 1 \leq u \leq T,~\inf_{z\in\mathds{R}}\{z+\dfrac{1}{u(1-\tau)}\sum_{s=1}^{u}(Y_{w}(\zeta_{w}^s)-z)^+\}\geq\cvar(Y_{x_{w}})+\tilde{\epsilon}_u^\eta\Big)\nonumber\\
		\leq~~ &\sum_{w=1}^{T}\sum_{u=1}^{T}\mathds{P}\Big(\inf_{z\in\mathds{R}}\{z+\dfrac{1}{u(1-\tau)}\sum_{s=1}^{u}(Y_{w}(\zeta_{w}^s)-z)^+\}\geq\cvar(Y_{x_{w}})+\tilde{\epsilon}_u^\eta\Big)\nonumber.
		\end{align}
		Thus by Brown's inequality 
		\begin{equation}
		(\ref{ff})	< \sum_{w=1}^{T}\sum_{u=1}^{T}\exp(-2(\tau\tilde{\epsilon}_u^\eta/(b-a))^2u)\nonumber.
		\end{equation}
		Taking 
		$$\tilde{\epsilon}_u^\eta=\dfrac{(b-a)}{\tau}\sqrt{\dfrac{\log(2T^2/\eta)}{2u}}$$
		provides the first part, $i.e$ (\ref{ff})$ ~<\dfrac{\eta}{2}$.
		
		We use the same scheme of proof to bound (\ref{ss}), the only difference comes from the fact that the inequality of deviation is different:
		\begin{align}&
		\mathds{P}\bigg(\forall h\geq 0, \forall 0\leq j \leq K^h,\nonumber \forall 1\leq t \leq T,~\widehat{\cvar}^{t}(Y_{x_{h,j}})\leq\cvar(Y_{x_{h,j}})-\epsilon_{N_{h,j}(t)}^\eta\bigg) \\
		\leq~~ &\mathds{P}\Big(\exists~ 1 \leq w \leq T,~\exists~ 1 \leq u \leq T,~\inf_{z\in\mathds{R}}\{z+\dfrac{1}{u(1-\tau)}\sum_{s=1}^{u}(Y_{w}(\zeta_{w}^s)-z)^+\}\leq\cvar(Y_{x_{w}})-\epsilon_u^\eta\Big)\nonumber\\
		\leq~~ &\sum_{w=1}^{T}\sum_{u=1}^{T}\mathds{P}\Big(\inf_{z\in\mathds{R}}\{z+\dfrac{1}{u(1-\tau)}\sum_{s=1}^{u}(Y_{w}(\zeta_{w}^s)-z)^+\}\leq\cvar(x_{w})-\epsilon_u^\eta\Big)\nonumber.
		\end{align}
		By Brown's inequality 
		\begin{equation}
		(\ref{ss})	< \sum_{w=1}^{T}\sum_{u=1}^{T}3\exp\bigg(-\dfrac{\tau}{5}\Big(\dfrac{\epsilon_u^\eta}{b-a}\Big)^2u\bigg)\nonumber
		\end{equation}
		Taking 
		$$\tilde{\epsilon}_u^\eta=(b-a)\sqrt{\dfrac{5\log(6T^2/\eta)}{\tau u}}$$
		provides (\ref{ss}) $<\dfrac{\eta}{2}$. 
		
		Finally putting (\ref{ff}) and (\ref{ss}) together provides $\mathds{P}\big(\xi_\eta\big)<\eta$ and hence $\mathds{P}(\xi_{\eta}^c)~=~\mathds{P}(\mathcal{A}_\eta)=1-\eta$.

			\textbf{Proof of Proposition} \ref{lern-mil}
				If $Y_1\cdots,Y_n$ are i.i.d random variables bounded by $(a,b)$ then Thomas-Learned-Miller's inequalities provide
				$$\mathds{P}\Bigg(-\cvar<\dfrac{1}{1-\tau}\sum_{i=1}^n(Y_{i+1}-Y_i)\Big(\dfrac{i}{n}-\sqrt{\dfrac{\log(1/\eta)}{2n}}-\tau\Big)^+-Y_{n+1}\Bigg)< \eta$$
				and
				$$\mathds{P}\Bigg(-\cvar>\dfrac{1}{1-\tau} \sum_{i=0}^{n-1}(Y_{i+1}-Y_i)\Big(\min\big\{1,\dfrac{i}{n}+\sqrt{\dfrac{\log(2T^2/\eta)}{2N_{h,j}(t)}}\big\}-\tau\Big)^+-Y_n\Bigg)<\eta.$$
				
				Define 	
				\begin{eqnarray*}
					\xi_{\eta,1}&=&\{\forall~ h\geq 0, \forall~ 0\leq j \leq K^h,\nonumber \forall~ 1\leq t \leq T, -\cvar(Y_{h,j})<U_{N_{h,j}(t)}^\eta\},
				\end{eqnarray*}
				and
				\begin{eqnarray*}
					\xi_{\eta,2}&=&\{\forall~ h\geq 0, \forall~ 0\leq j \leq K^h,\nonumber \forall~ 1\leq t \leq T, -\cvar(Y_{h,j}) > L_{N_{h,j}(t)}^\eta\},
				\end{eqnarray*}
				To treat the sequential point of view, here we use a double union bound as it is done in the proof of Proposition 13, then it can be shown that
				$$	\mathds{P}(\xi_{\eta,1})< \sum_{w=1}^{T}\sum_{u=1}^{T}\mathds{P}\Big(-\cvar(Y_w^u)<U_{u}^\eta\Big).$$
				Thus by defining
				$$U_{u}^\eta=\dfrac{1}{1-\tau} \sum_{i=0}^{u-1}(Y_{i+1}-Y_i)\Big(\min\big\{1,\dfrac{i}{u}+\sqrt{\dfrac{\log(2T^2/\eta)}{2u}}\big\}-\tau\Big)^+-Y_u$$
				we obtain
				$$	\mathds{P}(\xi_{\eta,1})< \sum_{w=1}^{T}\sum_{u=1}^{T}\dfrac{\eta}{2T^2}=\dfrac{\eta}{2}.$$
				Using the same scheme of proof with 
				$$L_{u}^\eta= \dfrac{1}{1-\tau}\sum_{i=1}^u(Y_{i+1}-Y_i)\Big(\dfrac{i}{u}-\sqrt{\dfrac{\log(2T^2/\eta)}{2u}}-\tau\Big)^+-Y_{u+1}$$
				provides 
				$$	\mathds{P}(\xi_{\eta,2})<\dfrac{\eta}{2}.$$
				Finally 
				$$\mathds{P}(\xi_{\eta,1}\cup \xi_{\eta,1})<\eta,$$
				and hence $\mathds{P}\Big((\xi_{\eta,1}\cup \xi_{\eta,1})^c\Big)=\mathds{P}(\mathcal{A}_\eta)=1-\eta$.
		
\end{document}